%% file: main.tex
\documentclass[10pt,twocolumn,letterpaper]{article}

\usepackage[pagenumbers]{cvpr} 

\input{preamble}
\definecolor{cvprblue}{rgb}{0.21,0.49,0.74}
\usepackage[pagebackref,breaklinks,colorlinks,allcolors=cvprblue]{hyperref}

\def\confName{CVPR}
\def\confYear{2027}

\title{RoG-DAgger: Rollout-Guided Post-Training for End-to-End Driving}

\author{
    Liangyu Zhong$^{1,2}$ \quad
    Joachim Sicking$^{1}$ \quad
    Fabian H\"uger$^{1}$ \quad
    Hanno Gottschalk$^{2}$\\[4pt]
    $^{1}$CARIAD SE, Volkswagen Group, Germany\\
    $^{2}$Institute of Mathematics, Technical University of Berlin, Germany
}

\begin{document}
\maketitle
\input{sec/0_abstract}    
\input{sec/1_intro}
\input{sec/2_related_work}
\input{sec/3_rog_dagger}
\input{sec/4_experiments}

\input{sec/5_discussion}
\vspace{-8mm}
{
    \small
    \bibliographystyle{ieeenat_fullname}
    \bibliography{main}
}



\clearpage
\input{sec/X_suppl}


\end{document}

%% file: preamble.tex
\usepackage{algorithm}
\usepackage{algorithmic}
\usepackage{makecell}
\usepackage{enumitem}
\usepackage{xurl}
\usepackage{multirow}
\usepackage{xcolor}
\usepackage[table]{xcolor}
\usepackage{siunitx}
\usepackage{cuted}
\usepackage{capt-of} 

%% file: sec/0_abstract.tex
\begin{abstract}
Recent end-to-end driving systems demonstrate strong performance on closed-loop benchmarks, yet are still predominantly trained on fixed expert-collected data using open-loop imitation learning. This training-inference mismatch leaves the policy vulnerable in policy-induced states, where accumulated errors can lead to safety-critical failures. 
Dataset Aggregation (DAgger) mitigates this issue
by post-training policies on expert demonstrations
collected at policy-induced states during closed-loop execution.
Existing driving \mbox{DAgger} pipelines, however, face three challenges: restricted expert trajectory-and-speed solutions, premature or delayed takeover, and expert supervision that relies on information the student cannot observe.
To address this, we introduce \mbox{\textbf{RoG-DAgger}}, a post-training framework that uses short-horizon kinematic rollouts to guide both expert supervision and takeover timing. Specifically,
RoG-DAgger expands the expert's trajectory-and-speed solution space and evaluates candidate plans through rollout to construct preventive supervision. 
Moreover, it uses rollout solvability to guide takeover timing, balancing premature intervention against the risk of intervening too late for effective correction.
Lastly, it aligns the expert's field of view with that of the student to provide student-compatible supervision. 
Across in-distribution (including long-horizon) and out-of-distribution evaluations, RoG-DAgger outperforms the end-to-end model SimLingo, achieving a 5.0 points higher driving score and a 6.0 p.p.\ higher success rate on Bench2Drive, a driving score of 45 versus 22 on Longest6 v2, and an out-of-distribution success rate of 63\% versus 55\% on Fail2Drive.
\end{abstract}

%% file: sec/1_intro.tex
\section{Introduction}
\label{intro}

\begin{figure}[bt]
\centering
\includegraphics[width=0.98\columnwidth]{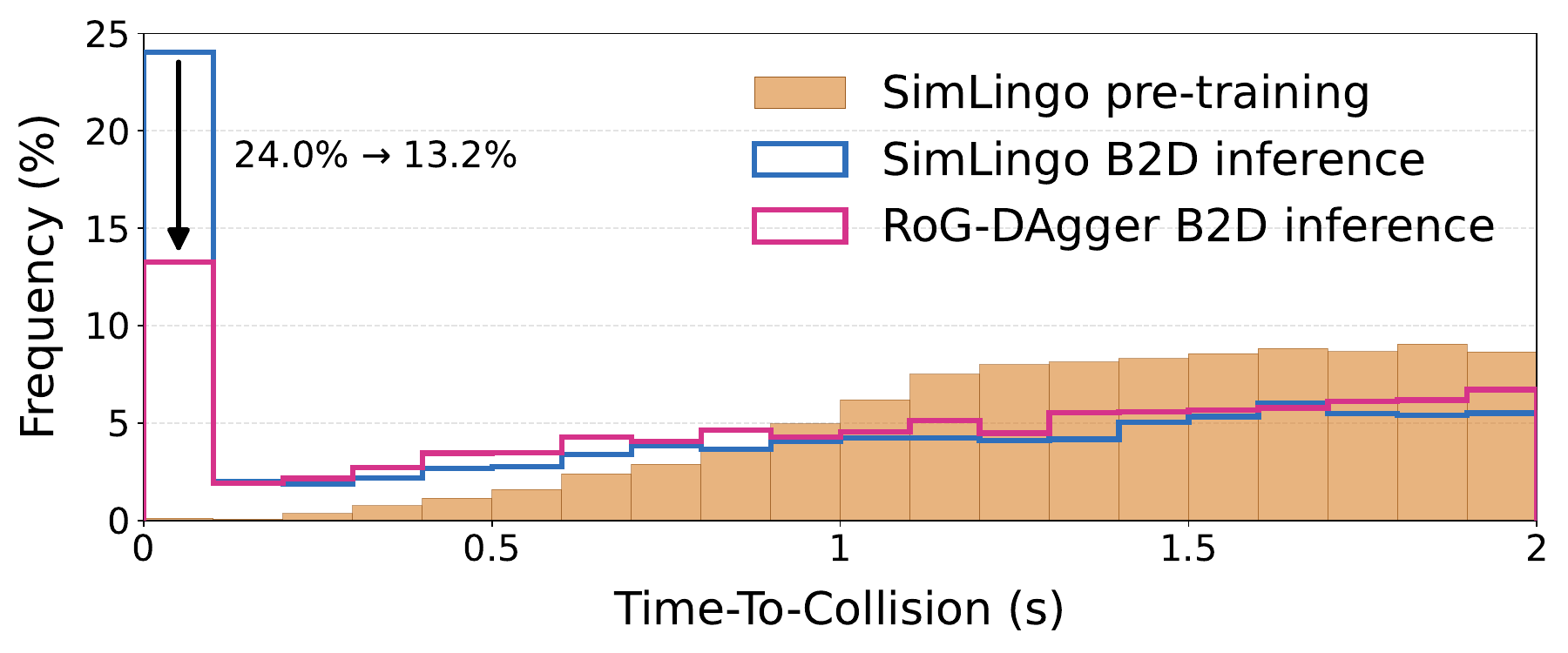}
\caption{
\textbf{SimLingo pre-training data (orange) lacks safety-critical interactions as measured by time-to-collision (TTC), limiting the model's ability to learn how to prevent or recover from them.} In closed-loop Bench2Drive inference (blue), however, such interactions occur frequently, which constitutes a substantial distribution shift. RoG-DAgger post-training (magenta) reduces the fraction of near-collision samples (TTC~$\leq0.1$~s) by 45\%. Frequencies are normalized over samples with TTC~$\leq2$~s.
}

\label{ttc_compare}
\end{figure}

Recent end-to-end (E2E) driving systems~\cite{simlingo, orion, LEAD} have achieved strong results on closed-loop benchmarks, where predicted actions are executed in the simulator and influence subsequent observations. However, these models are still predominantly trained on fixed expert-collected data using open-loop imitation learning. They therefore remain vulnerable to the distribution shift between open-loop training and closed-loop execution~\cite{takead}. We study this issue using SimLingo~\cite{simlingo}, a widely used camera-based E2E driving model. Evaluating SimLingo on the closed-loop Bench2Drive benchmark (B2D)~\cite{b2d}, we find that collisions account for roughly 80\% of its failed routes (see App.~\ref{model_details} for details). These failures likely arise when small imitation errors propagate and compound over time, causing the policy to deviate from expert trajectories and visit states underrepresented in the training data, often referred to as compounding errors~\cite{compounding}. To illustrate the resulting train–inference mismatch, we compare how frequently the model encounters safety-critical interactions, measured by the time-to-collision (TTC)~\cite{nuplan} distributions of SimLingo's training data and its inference on B2D. As shown in Fig.~\ref{ttc_compare}, the SimLingo driving model encounters substantially lower TTC values during closed-loop inference than observed during training. This indicates that the model may learn neither how to \textit{prevent} such near-collision events nor how to \textit{recover} from collisions once they occur.

\begin{figure*}[t]
\centering
\includegraphics[width=2\columnwidth]{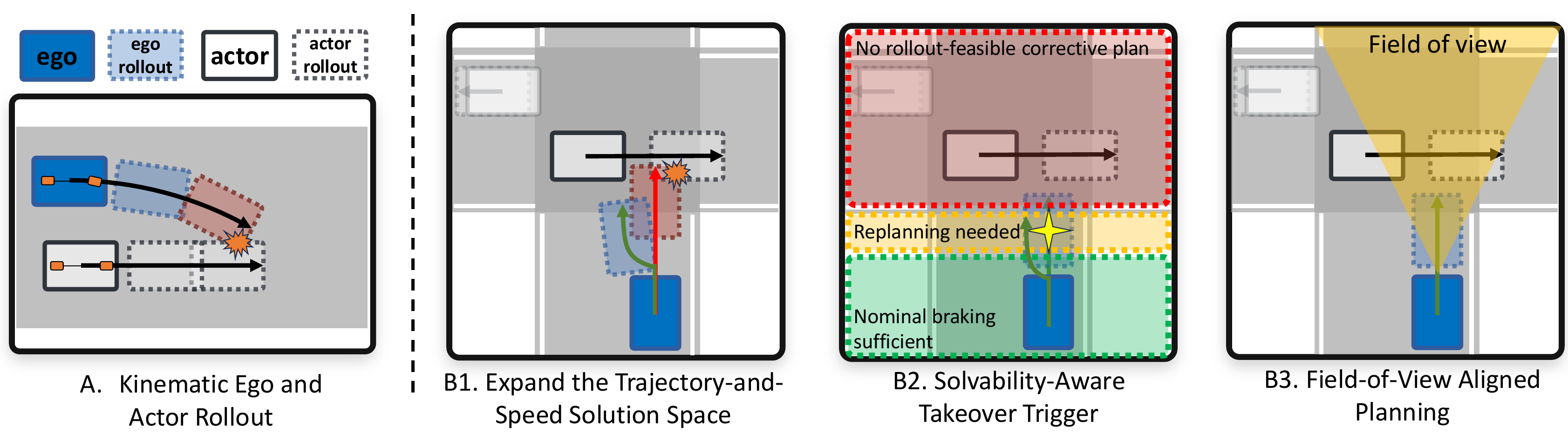}
\caption{
\textbf{RoG-DAgger leverages kinematic rollouts
to improve expert supervision and takeover timing.}
(A) Kinematic rollouts evaluate candidate
trajectory-and-speed plans to construct preventive
expert supervision. Three improvements to DAgger:
(B1) Expanded trajectory-and-speed solution
space (red: collision-prone nominal plan;
green: corrective plan).
(B2) Solvability-aware trigger:
nominal braking is sufficient (green), corrective
replanning is required and feasible (yellow),
or no rollout-feasible corrective plan is found
(red). The yellow star marks expert takeover.
(B3) Expert-student field-of-view alignment
(yellow triangle) for student-learnable supervision.
}
\label{teaser}
\end{figure*}

\textbf{D}ataset \textbf{Agg}regation (DAgger)~\cite{dagger} provides a promising way to mitigate this issue by running an expert model alongside a pretrained E2E policy in closed-loop simulation and fine-tuning the policy using the expert’s demonstrations. Here, the expert is a generally stronger privileged driving policy with access to scene information, such as surrounding-actor states and lane geometry. In driving DAgger, expert supervision is typically collected around \emph{takeover events}, where control is transferred from the student policy to the expert. Recent methods such as TakeAD~\cite{takead} primarily collect post-takeover expert demonstrations, providing supervision for recovery from unsafe states. TakeVLA~\cite{takevla} additionally retains pre-takeover frames and assigns warning-style language labels (e.g., ``decelerate'') to encourage proactive avoidance, while further refining low-level behavior through reinforcement learning. However, TakeVLA's pre-takeover labels remain event-level rather than state-specific, even though different states associated with the same warning may require substantially different trajectories and speed profiles. We therefore seek to provide explicit trajectory-and-speed supervision for each pre-takeover state by constructing counterfactual rollouts of expert-proposed candidate plans.
Generating reliable preventive supervision, however, remains challenging due to three limitations of existing DAgger pipelines: i) a restricted expert trajectory-and-speed solution space, ii) takeover timing that may be premature or too late to prevent the impending failure, and iii) a misalignment between the fields of view of expert \mbox{and student}.

\begin{figure*}[!t]
\centering
\includegraphics[width=1.6\columnwidth]{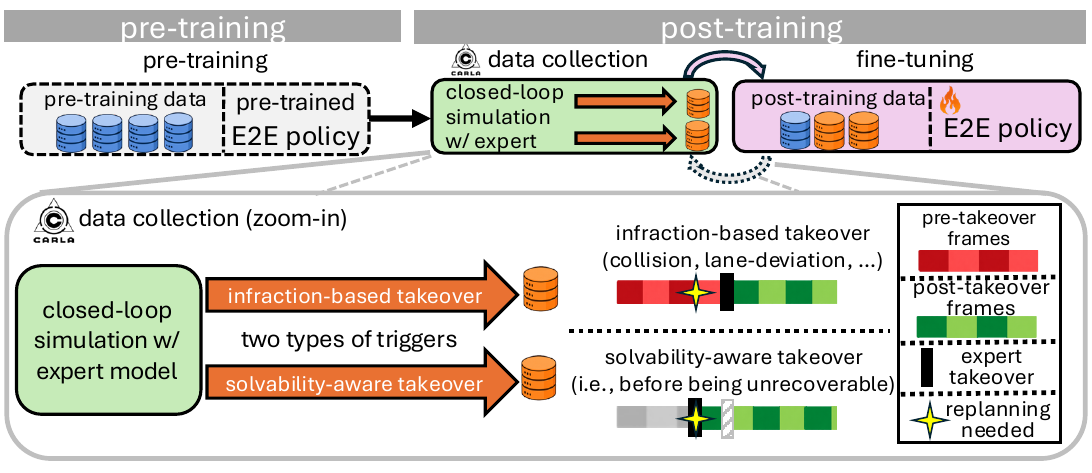}
\caption{\textbf{Overview of the RoG-DAgger training pipeline.} Starting from a pretrained E2E policy, we collect post-training data in two independent closed-loop runs (see zoom-in panel) and fine-tune the policy using a mixture of pre- and post-training data. The infraction-based run captures policy-induced failures and expert recovery demonstrations, while the solvability-aware run guides preventive takeover timing and retains subsequent expert demonstrations. (gray hatched bar indicates avoided collision). \mbox{A restart} trigger is omitted for clarity.}
\label{overall}
\end{figure*}

To address these limitations, we propose RoG-DAgger, a \textbf{Ro}llout-\textbf{G}uided post-training framework for E2E driving (Fig.~\ref{overall}). At its core, kinematic rollouts enable the evaluation of candidate expert plans, guiding both preventive
supervision and takeover timing (Fig.~\ref{teaser}, A). To address the first limitation, we expand the expert's trajectory-and-speed solution space with multiple candidate geometric trajectories and target speeds, and retain only rollout-solvable candidates that satisfy driving constraints (Fig.~\ref{teaser}, B1). To address the second, we introduce a solvability-aware trigger that uses rollout evaluation to guide the timing of preventive expert takeover (Fig.~\ref{teaser}, B2). Finally, to address the third limitation, we align the fields of view of expert and student to reduce supervision that depends on information unavailable to the student, similar to \cite{LEAD}, see Fig.~\ref{teaser}, B3.

We evaluate RoG-DAgger by post-training SimLingo on the public CARLA Leaderboard 2.0 training routes~\cite{carla_lb2} and testing on three complementary closed-loop benchmarks: B2D for in-distribution evaluation, Longest6 v2 for long-horizon driving, and Fail2Drive for out-of-distribution generalization. On B2D, RoG-DAgger outperforms SimLingo by 5.0 driving score (DS) points and 6.0 p.p.~in success rate (SR), surpassing TakeAD and achieving performance comparable to TakeVLA. It further doubles SimLingo's DS from 22 to 45 on Longest6 v2 and improves OOD SR from 55\% to 63\% on Fail2Drive.

Overall, our contributions are three-fold:
\begin{itemize}
\item We develop a controller-aligned kinematic rollout
mechanism as the foundation for evaluating expert
trajectory-and-speed plans, enabling state-specific preventive
supervision and solvability-aware takeover.

\item Building on this mechanism, we propose RoG-DAgger, a post-training framework that improves both expert supervision and takeover by expanding the expert solution space, guiding the timing
of preventive takeover, and aligning the fields of view of expert and student.

\item We demonstrate that RoG-DAgger improves closed-loop driving across complementary in-distribution (including long-horizon) and out-of-distribution settings, showing that rollout-guided supervision generalizes beyond the scenarios used for post-training.
\end{itemize}

%% file: sec/2_related_work.tex
\section{Related Work}\label{sec:related_work}
We first review representative end-to-end driving systems in Sec.~\ref{e2e_sys}. Online post-training approaches that enable these systems to perform better in closed-loop settings are discussed in Sec.~\ref{online_post_train}.
\subsection{End-to-End Driving Systems}\label{e2e_sys}
End-to-end driving systems directly optimize the mapping from sensory observations to planned trajectories, enabling a fully differentiable pipeline that reduces reliance on hand-crafted interfaces between different driving modules. Earlier (non-language-based) approaches include TransFuser~\cite{transfuser} and UniAD~\cite{uniad}, which jointly optimize perception, prediction, and planning tasks within a planning-oriented framework.

Recently, vision-language models have increasingly been adapted to end-to-end driving, due to their improved scene understanding and broad world knowledge~\cite{drivelm,vl4ad,focus, braking_llm}, giving rise to vision-language-action models. Representative methods include SimLingo~\cite{simlingo}, ORION~\cite{orion}, and AutoVLA~\cite{autovla} which have primarily focused on improving the alignment between visual observations, language instructions, and driving actions. However, despite the growing emphasis on closed-loop simulation in the community, these models are still primarily trained on fixed offline datasets and do not explicitly address the distribution shift between open-loop and closed-loop (see introduction and~\cite{takead}). This motivates closed-loop post-training methods that adapt pretrained driving policies using feedback or supervision derived from their behavior during deployment.
\subsection{Online Post-Training for E2E Systems}\label{online_post_train}
Unlike fixed offline training, \textit{online} post-training refines a pretrained driving policy using data collected through closed-loop interaction, where the policy’s actions influence subsequent states, observations, and the responses of surrounding actors. This enables the policy to learn from its own closed-loop errors. Existing approaches mainly differ in their supervision signal. Reward-based methods, such as MindDrive~\cite{minddrive}, optimize the policy using environment-derived feedback, e.g., traffic-infraction penalties. Expert-supervised methods commonly build on Dataset Aggregation (DAgger~\cite{dagger}), which queries a privileged expert at policy-induced states and uses expert demonstrations for further policy optimization.

Driving DAgger methods can be further characterized along two axes: \emph{when} expert takeover is triggered and \emph{which type of} expert supervision is collected. HG-DAgger~\cite{hg_dagger} relies on human experts to determine when to intervene and when to return control to the student, collecting expert recovery demonstrations for subsequent fine-tuning. ThriftyDAgger~\cite{ThriftyDAgger} automates intervention decisions using policy novelty, estimated through ensemble disagreement of different policies, and predicted task risk, estimated by a separately trained value network. DriveGPT4-V2~\cite{drivegptv2} employs an LLM-based privileged expert and triggers intervention when the discrepancy between student and expert predictions exceeds a predefined threshold. TakeAD~\cite{takead} collects expert demonstrations using predicted collision risk apart from expert-student discrepancies as takeover triggers, primarily providing post-takeover supervision. TakeVLA~\cite{takevla} additionally retains pre-takeover observations and assigns warning-style language labels to encourage preventive behavior.

Despite their effectiveness, these approaches have several limitations. Human-gated interventions (HG-DAgger) benefit from
expert judgment but are costly and difficult to scale.
Learned novelty and risk estimates (ThriftyDAgger) may become unreliable under distribution shifts, particularly in previously unseen safety-critical scenarios. Moreover, expert-student disagreement (DriveGPT4-V2, TakeAD and TakeVLA) does not
necessarily imply unsafe behavior, potentially
leading to interventions at non-critical states.
Existing takeover criteria also do not explicitly
assess corrective-plan feasibility. In terms of supervision, post-takeover demonstrations
provide corrective actions but omit direct action
supervision at preceding policy-induced states,
while warning-style language labels do not explicitly
specify the required corrective actions.
RoG-DAgger addresses these limitations by constructing
state-specific trajectory-and-speed supervision for
pre-takeover states and guiding takeover timing
through rollout-based assessment of nominal braking
sufficiency and corrective-plan feasibility.
This aims to preserve student exposure to informative
safety-critical states while retaining sufficient
opportunity for expert intervention, facilitating
targeted data aggregation and policy improvement.

%% file: sec/3_rog_dagger.tex
\section{Rollout-Guided DAgger for E2E Driving}\label{para:rog_dagger}
We first formulate the learning setup and overall RoG-DAgger framework in Sec.~\ref{problem} and Sec. \ref{rog_dagger_framework}, respectively. We then introduce the kinematic rollout model used to evaluate candidate trajectory-and-speed plans in Sec.~\ref{kinematic_rollout}, followed by the rollout-guided expert supervision and takeover mechanisms in Sec.~\ref{rog_expert}.
\subsection{Problem Formulation}\label{problem}
Driving policies in E2E driving map the current sensory observation, ego state, and navigational condition at time step $t$ to a future ego plan
\begin{equation}
\hat{\mathbf a}_t
=
\Pi_\theta
\left(
\mathbf C_t,
\mathbf S_t^{\mathrm{ego}},
\mathbf O_t
\right),
\end{equation}
where $\Pi_\theta$ denotes the learned policy, $\mathbf C_t$ the navigational condition (e.g., GPS target points or language instructions), $\mathbf S_t^{\mathrm{ego}}$ the ego state (i.e., position and speed), and $\mathbf O_t$ the sensory observation, such as RGB images. Following SimLingo, we represent the predicted ego plan as
\begin{equation}
\hat{\mathbf a}_t
=
\left(
\hat{\mathbf a}^{\mathrm{geo}}_t,
\hat{\mathbf a}^{\mathrm{temp}}_t
\right),
\end{equation}
where $\hat{\mathbf a}^{\mathrm{geo}}_t$ denotes the geometric trajectory, represented by a sequence of equidistantly sampled waypoints describing the intended path, and $\hat{\mathbf a}^{\mathrm{temp}}_t$ specifies the ego vehicle's temporal progress along this path. During execution, the predicted trajectories are converted into steering, throttle, and braking commands using lateral and longitudinal PID controllers~\cite{pid}.

Driving policies are commonly trained on expert demonstrations using an L1 imitation loss (IL)
\begin{equation} \label{eq:il_loss}
\mathcal{L}_{\mathrm{IL}} =
\left\|
\hat{\mathbf a}^{\mathrm{geo}}
-
\mathbf a^{\mathrm{geo,}}{^{*}}
\right\|_1
+
\left\|
\hat{\mathbf a}^{\mathrm{temp}}
-
\mathbf a^{\mathrm{temp},}{^{*}}
\right\|_1,
\end{equation}
where $\mathbf a^{*}$ denotes the expert supervision.
During pre-training, this supervision is collected at expert-induced states, whereas DAgger~\cite{dagger} queries the expert at states visited by the student policy, providing corrective supervision for policy-induced states. Following recent driving DAgger methods~\cite{takead,takevla}, we selectively collect data around takeover events rather than at every policy step. We refer to states before and after takeover as \emph{pre-} and \emph{post-takeover} samples, respectively. Post-takeover trajectory supervision is directly constructed from the expert-controlled rollout, while our goal is to additionally construct state-specific trajectory-and-speed supervision for pre-takeover states.
\subsection{RoG-DAgger Framework}
\label{rog_dagger_framework}
We now provide an overview of our RoG-DAgger framework (Fig.~\ref{overall}).
Starting from a pretrained student policy, we perform two independent closed-loop data-collection runs using \emph{infraction-based} and \emph{solvability-aware} takeover triggers, respectively. Running the two protocols separately is necessary because solvability-aware takeover changes the subsequent state distribution and may suppress failures that the student would otherwise encounter.

In the infraction-based run, takeover is triggered by collision, lane deviation, red-light or stop-sign violation. We retain the pre-takeover states leading to the failure together with post-takeover expert recovery demonstrations. In the solvability-aware run, in contrast, takeover occurs before an impending failure according to a rollout-based criterion (details in Sec.~\ref{early_takeover}) and only post-takeover expert demonstrations are retained. In both runs, moreover, a restart trigger, similar in purpose to that used in TakeVLA~\cite{takevla}, is enabled when the ego vehicle fails to make expected progress for a prolonged period, with only post-takeover recovery samples being retained.
Detailed trigger definitions are provided in App.~\ref{trigger_condition}. The collected samples are subsequently used for fine-tuning with the imitation objective mentioned above, which yields the RoG-DAgger post-\mbox{trained model}.

In the following, we describe RoG-DAgger in greater detail, beginning with
the kinematic rollout model used to construct preventive supervision and to assess \mbox{candidate solvability}.

\subsection{Kinematic Ego and Actor Rollout}\label{kinematic_rollout}
RoG-DAgger uses short-horizon rollouts of the ego vehicle and surrounding actors to evaluate candidate trajectory-and-speed plans. These rollouts support both preventive supervision for pre-takeover states and the solvability-aware takeover mechanism. We first describe the ego rollout, followed by the surrounding-actor rollout. Given the current ego state $\mathbf{s}_t^{\mathrm{ego}}$ (including speed $v_t$), a candidate geometric trajectory $\mathbf{a}_t^{\mathrm{geo}}$, and target speed $v_t^{\mathrm{target}}$, we define the ego rollout as
\begin{equation}
\mathbf{a}_t^{\mathrm{rollout}}
=
\mathcal{R}_{\mathrm{ego}}
\left(
\mathbf{s}_t^{\mathrm{ego}},
\mathbf{a}_t^{\mathrm{geo}},
v_t^{\mathrm{target}}
\right),
\end{equation}
which represents the resulting counterfactual ego trajectory. Our rollout closely reflects the execution of candidate trajectory-and-speed plans. At each step, we apply the same lateral and longitudinal control logic used during deployment and propagate the resulting controls using the kinematic bicycle model and dynamics parameters from World on Rails~\cite{worldonrail}. Crucially, the rollout reproduces the deployed braking condition, enabling realistic transitions between throttle and full braking. Unlike PDM-Lite~\cite{pdm_lite}, which models throttle control without this braking behavior, our rollout more faithfully reflects the deployed longitudinal controller. For surrounding actors, we follow the non-reactive rollout setting of PDM-Lite, propagating each actor over the same horizon while assuming unchanged low-level control. Together, the ego and actor rollouts provide the counterfactual future trajectories that are used to evaluate candidate trajectory-and-speed plans, as described in the following sections. Further details on the kinematic ego rollout, including its pseudocode, are provided in App.~\ref{app:kinematic_ego_rollout}.

\subsection{Rollout-Guided Expert Supervision}\label{rog_expert}
We now use the rollout model from Sec.~\ref{kinematic_rollout} and examine whether expert supervision assigned to pre-takeover states is actually preventive. In a diagnostic run, we use CARLA collision infractions as takeover triggers and PDM-Lite~\cite{pdm_lite}, a commonly used privileged expert with near-perfect performance on B2D and employed by prior driving DAgger work~\cite{takead,takevla}. For each pre-takeover state, we roll out the corresponding expert trajectory-and-speed plan and assess whether it prevents the \mbox{impending collision}.

Our analysis shows that a non-negligible fraction of pre-takeover samples remain \textit{collision-prone} under rollout (additional details are provided in App.~\ref{pre_pdm_dagger}). This suggests that expert supervision at pre-takeover states is not necessarily sufficient to prevent the impending collision. 
We consider two potential limitations related to prevention feasibility: the expert's restricted trajectory-and-speed solution space and takeover timing that may be too late for effective collision avoidance. Separately, the mismatch between privileged expert inputs and the student's field of view may hinder learning from expert supervision. RoG-DAgger addresses these three challenges by expanding the expert's solution space, introducing solvability-aware takeover, and aligning the expert's field of view with that of the student. We elaborate on each aspect in the following.

\subsubsection{Expanding the Trajectory-and-Speed Solution Space}
PDM-Lite follows a precomputed route and responds to predicted collisions with dynamic actors primarily by setting the target speed to zero, resulting in full braking along the nominal path. This is effective when longitudinal control suffices, but does not verify that braking prevents the collision and may fail when geometric adaptation is required (see examples in App.~\ref{para:examples_geo_shifting}). We therefore expand the expert's trajectory-and-speed solution space and validate candidate plans through the rollout model.

When PDM-Lite predicts a collision requiring braking, we generate geometric candidates by laterally shifting the nominal trajectory $\mathbf a_t^{\mathrm{geo}}$. We first identify the waypoints corresponding to the predicted overlap between the ego and actor rollouts. We then shift these waypoints laterally, away from the actor's direction of motion. We increase the offset in steps of $\Delta d$ up to $d_{\max}$, using cosine ease-in/ease-out transitions, and discard trajectories leaving the admissible lanes. For each geometric candidate $\mathbf a_{t,i}^{\mathrm{geo}}$, indexed by $i$, we evaluate a set of target speeds indexed by $j$. Outside junctions, we retain zero target speed. Inside junctions, where stopping in the conflict region may be undesirable, we additionally consider non-zero evasive speeds
\begin{equation}
    \mathcal V_\mathrm{target} =
\begin{cases}
\{0,\;0.25\;v_t,\;0.5\;v_t\}, & \text{in junctions},\\
\{0\}, & \text{otherwise},
\end{cases}
\end{equation}
and compute the corresponding ego rollout
$
\mathbf a_{t,i,j}^{\mathrm{rollout}}
=
\mathcal R_{\mathrm{ego}}
\left(
\mathbf S_t^{\mathrm{ego}},
\mathbf a_{t,i}^{\mathrm{geo}},
v_{t,j}^{\mathrm{target}}
\right).$ We retain only collision-free, lane-compliant candidates, prioritizing higher target speed and then smaller lateral offset. If no valid candidate exists, we use the nominal trajectory with full braking as a best-effort fallback. For each retained pre-takeover observation, the selected expert trajectory and its kinematic rollout are converted into the geometric and temporal waypoint targets used in Eq.~\ref{eq:il_loss}. Additional details regarding solution space expansion are provided in App.~\ref{traj_and_speed_expansion}. 

\subsubsection{Solvability-Aware Takeover Trigger} \label{early_takeover}
A richer expert solution space alone cannot ensure
effective preventive supervision if takeover occurs
at an inappropriate time. Premature takeover limits
the student's exposure to safety-critical
policy-induced states, whereas delayed intervention
may leave insufficient opportunity for successful
correction. We therefore seek a takeover timing
that balances exposure to critical states with
the remaining feasibility of corrective action.

Existing methods trigger takeover based on predicted collision risk or expert-student disagreement. TakeAD~\cite{takead} uses predicted collision risk and steering discrepancies, while TakeVLA~\cite{takevla} additionally considers insufficient student deceleration and constant-velocity collision prediction. However, these criteria do not explicitly assess whether corrective intervention is necessary and still feasible. In particular, constant-velocity prediction may trigger prematurely by neglecting possible braking and steering responses.

We introduce a solvability-aware trigger that uses
kinematic rollouts to guide takeover timing.
The goal is to intervene when nominal emergency
braking becomes insufficient, while corrective
maneuvers remain feasible. Specifically, when PDM-Lite predicts a collision
and requests emergency braking, we evaluate
the nominal braking plan using our rollout model.
If this plan is predicted to avoid the collision, the student
remains in control. Otherwise, we evaluate
corrective trajectory-and-speed plans from the expert's expanded solution space.
When a collision-free solution exists, takeover
is triggered either under imminent collision risk
(\texttt{Replan Success}), or after repeated
confirmation for less urgent conflicts
(\texttt{Repeated \mbox{Replan Success}}).

These conditions guide takeover toward critical
policy-induced states where corrective intervention
is both necessary and feasible. When no rollout-feasible candidate is found under imminent collision risk
(\texttt{Replan Failure}), we trigger emergency
braking as a fallback. Detailed trigger conditions
and examples are provided in
Tab.~\ref{tab:collection_trigger_conditions}
and App.~\ref{para:solvability_aware_examples}, while the effectiveness of the solvability-aware trigger is evaluated in
App.~\ref{app:solvability_aware_timing}.

\subsubsection{Field-of-View-Aligned Planning}
Even a rollout-solvable expert plan may provide ineffective supervision if it depends on information unavailable to the student. Following~\cite{LEAD}, we therefore restrict expert planning to surrounding actors within the student's field of view.

DAgger introduces an additional issue when an infraction is caused by an actor outside the student's field of view. In such cases, the observation-aligned expert also cannot condition its action on the responsible actor, so its pre-takeover trajectory does not provide meaningful supervision for preventing that event. For example, during a lane change, the vehicle responsible for a later collision may still be outside the student's field of view in early pre-takeover states. We therefore discard pre-takeover samples whose responsible actor is not visible to the student, while retaining the associated post-takeover samples for recovery supervision. We refer to the resulting expert as RoG-Expert.

%% file: sec/4_experiments.tex
\section{Experiments}
We first introduce our experimental setup and implementation details in Sec.~\ref{para:main_imp_details}. We then present the performance of RoG-DAgger on in-distribution and out-of-distribution benchmarks in Secs.~\ref{id_benchmark} and~\ref{ood_benchmark}, respectively. Finally, we validate our design
choices through ablation studies in
Sec.~\ref{para:ablation_studies} and analyze
the effects of expert choice and takeover timing
in Sec.~\ref{sec:trigger_timing}.
\subsection{Implementation Details}\label{para:main_imp_details}
\begin{table}[t]
\centering

{\small
\setlength{\tabcolsep}{1mm}
\caption{\textbf{Evaluation on Bench2Drive and Longest6 v2.}
DS, SR, and RC denote Driving Score, Success Rate, and Route Completion, respectively. RoG-DAgger achieves strong performance across both benchmarks. -- indicates implementation is unavailable. $^{*}$ represents evaluation using official checkpoints. TakeVLA-noRL indicates a TakeVLA-style baseline using its triggers and expert (PDM-Lite) but no RL.}
\label{tab:b2d_longest6}
\begin{tabular}{@{}lccrr@{}}
\toprule
\multirow{2}{*}{\textbf{Method}} &
\multicolumn{2}{c}{\textbf{Bench2Drive}} &
\multicolumn{2}{c}{\textbf{Longest6 v2}} \\
\cmidrule(lr){2-3}
\cmidrule(lr){4-5}
&
\textbf{DS} $\uparrow$ &
\textbf{SR} $\uparrow$ &
\textbf{DS} $\uparrow$ &
\textbf{RC} $\uparrow$ \\
\midrule

\multicolumn{5}{l}{\hspace{1em}\textit{Privileged methods}} \\
\midrule

PDM-Lite~\cite{pdm_lite}
& 97.02 & 92.27
& 73 & 100 \\

\rowcolor{gray!15}
RoG-Expert~\textbf{(ours)}
& 93.41 & 81.81
& 55 & 100 \\
\midrule

\multicolumn{5}{l}{\hspace{1em}\textit{Pre-trained-only methods}} \\
\midrule

UniAD~\cite{uniad}
& 45.81 & 16.36
& 5 & 9 \\

TF$++$~\cite{pdm_lite}
& 84.21 & 67.27
& 23 & 70 \\

ORION~\cite{orion}
& 77.74 & 54.62
& ${}^*7$ & ${}^*29$ \\

AutoVLA~\cite{autovla}
& 78.84 & 57.73
& -- & -- \\

HiP-AD~\cite{hipad}
& 86.77 & 69.09
& 7 & 56 \\

SimLingo~\cite{simlingo}
& $85.07\,{\scriptstyle\pm 0.95}$
& $67.27\,{\scriptstyle\pm 2.11}$
& $22\,{\scriptstyle\pm 4}$
& $70\,{\scriptstyle\pm 8}$ \\

\midrule
\multicolumn{5}{l}{\hspace{1em}\textit{Post-trained methods}} \\
\midrule

MindDrive~\cite{minddrive}
& 78.04 & 55.09
& ${}^*6$ & ${}^*21$ \\

TakeAD~\cite{takead}
& 71.39 & 40.83
& -- & -- \\

TakeVLA~\cite{takevla}
& 89.72 & \textbf{73.73}
& -- & -- \\

TakeVLA-noRL
& $87.61\,{\scriptstyle\pm 0.26}$
& $67.38\,{\scriptstyle\pm 0.84}$
& $36\,{\scriptstyle\pm 1}$
& $\boldsymbol{83}\,{\scriptstyle\boldsymbol{\pm 2}}$ \\

\rowcolor{gray!15}
RoG-DAgger~\textbf{(ours)}
& $\boldsymbol{90.11}\,{\scriptstyle\boldsymbol{\pm 0.20}}$
& $73.29\,{\scriptstyle\pm 1.38}$
& $\boldsymbol{45}\,{\scriptstyle\boldsymbol{\pm 2}}$
& $\boldsymbol{83}\,{\scriptstyle\boldsymbol{\pm 4}}$ \\

\bottomrule
\end{tabular}
}

\end{table}

\begin{table}[t]
\centering

{\small
\setlength{\tabcolsep}{1mm}
\caption{\textbf{Evaluation on Fail2Drive.}
\textit{Base} denotes in-distribution evaluation, while \textit{Generalization} refers to distribution-shifted scenarios. DS, SR, and HM denote Driving Score, Success Rate, and their harmonic mean, respectively. RoG-DAgger remains competitive on base scenarios while showing improved OOD generalization \mbox{over SimLingo.}}
\label{tab:fail2drive}
\resizebox{\columnwidth}{!}{%
\begin{tabular}{@{}lcc>{\columncolor{gray!15}}ccc>{\columncolor{gray!15}}c@{}}
\toprule
\multirow{2}{*}{\textbf{Method}} &
\multicolumn{3}{c}{\textbf{Base}} &
\multicolumn{3}{c}{\textbf{Generalization}} \\
\cmidrule(lr){2-4}
\cmidrule(lr){5-7}
&
\textbf{DS} $\uparrow$ &
\textbf{SR} $\uparrow$ &
\cellcolor{white}\textbf{HM} $\uparrow$ &
\textbf{DS} $\uparrow$ &
\textbf{SR} $\uparrow$ &
\cellcolor{white}\textbf{HM} $\uparrow$ \\
\midrule

UniAD~\cite{uniad}
& 47.5 & 36.3 & 41.2
& 44.0 & 27.6 & 33.9 \\

TF$++$~\cite{pdm_lite}
& 83.3 & 78.5 & 80.8
& \textbf{75.4} & 61.1 & 67.5 \\

ORION~\cite{orion}
& 53.0 & 52.0 & 52.5
& 51.2 & 46.0 & 48.5 \\

HiP-AD~\cite{hipad}
& 74.1 & 70.7 & 72.4
& 67.1 & 56.7 & 61.5 \\

SimLingo~\cite{simlingo}
& 82.6 & 79.3 & 80.9
& 71.7 & 55.0 & 62.2 \\

\midrule
TakeVLA-noRL
& $80.9\,{\scriptstyle\pm 0.9}$
& $72.3\,{\scriptstyle\pm 3.1}$
& $76.4\,{\scriptstyle\pm 2.1}$
& $73.5\,{\scriptstyle\pm 1.9}$
& $58.3\,{\scriptstyle\pm 2.5}$
& $65.0\,{\scriptstyle\pm 2.3}$ \\

RoG-DAgger~\textbf{(ours)}
& $\boldsymbol{83.5}\,{\scriptstyle\boldsymbol{\pm 1.4}}$
& $\boldsymbol{80.3}\,{\scriptstyle\boldsymbol{\pm 3.4}}$
& $\boldsymbol{81.9}\,{\scriptstyle\boldsymbol{\pm 2.5}}$
& $74.2\,{\scriptstyle\pm 0.6}$
& $\boldsymbol{63.0}\,{\scriptstyle\boldsymbol{\pm 2.6}}$
& $\boldsymbol{68.1}\,{\scriptstyle\boldsymbol{\pm 1.8}}$ \\

\bottomrule
\end{tabular}
}
}

\end{table}

\begin{figure*}[!t]
\centering
\includegraphics[width=2.0\columnwidth]{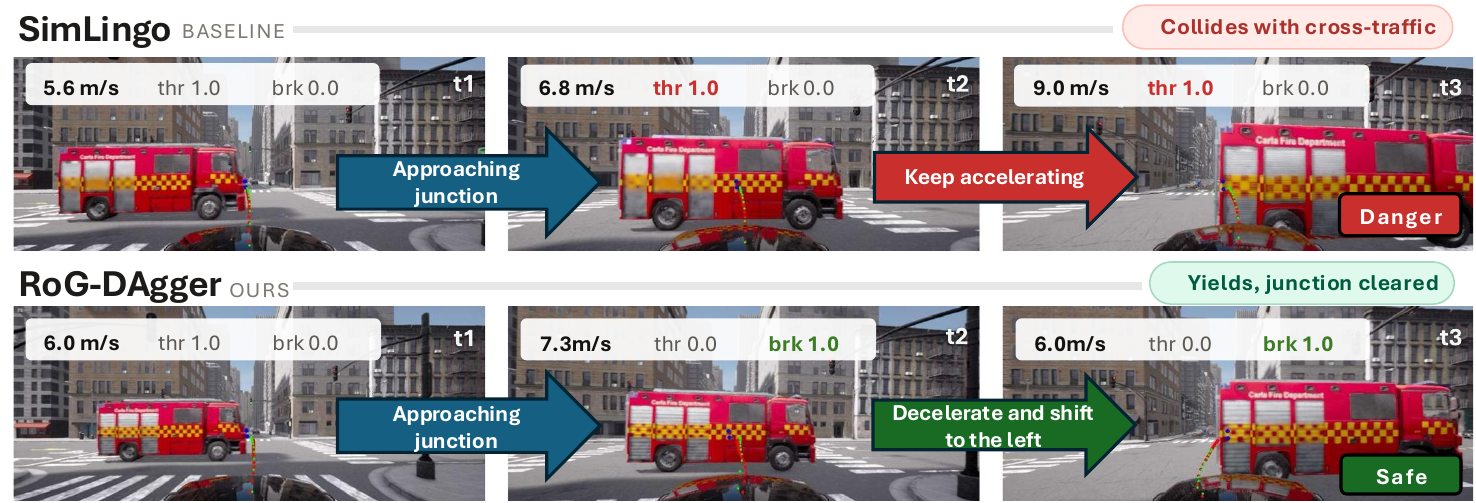}
\caption{\textbf{Qualitative comparison of RoG-DAgger with SimLingo on a Bench2Drive scene.} At $t_1$, both policies approach a junction while a cross-traffic fire truck is already visible. Both policies continue to accelerate, likely because the pre-training data contains primarily successful junction traversals and lacks corrective supervision for such hazardous interactions. As a result, SimLingo keeps accelerating at $t_2$ and collides with the fire truck at $t_3$. In contrast, after RoG-DAgger post-training, the policy begins braking at $t_2$ and shifts to the left, avoiding the collision at $t_3$. Red dots denote the predicted geometric trajectory, while green dots denote the predicted temporal trajectory. ``thr'': throttle, ``brk'': brake.}
\label{qua_example_1}
\end{figure*}
Following prior work~\cite{takevla}, we adopt SimLingo as the pretrained E2E driving policy. In main experiments, we replace the original InternVL-2 backbone~\cite{internvl} with the more recent Qwen3-VL-2B~\cite{qwen3}, which yielded comparable driving performance in our preliminary evaluation. To assess the generality of RoG-DAgger across VLM backbones, we additionally evaluate it with the original InternVL-2 backbone on B2D. Further details on the pre-training and backbone comparison are provided in App.~\ref{model_details}.

We then deploy the pretrained policy on the CARLA LB2 training routes~\cite{carla_lb2} and perform online post-training using RoG-DAgger as described in Sec.~\ref{para:rog_dagger}. Following~\cite{takead}, we reduce data-collection and fine-tuning costs by focusing on routes where the pretrained policy performs poorly and applying rule-based filtering (see App.~\ref{data_filtering}) to remove uninformative samples, such as cases where the ego vehicle remains stuck in traffic. Overall, we collect approximately 51K post-training samples and mix them with the original SimLingo pretraining data at a 7:3 post-to-pre ratio to mitigate catastrophic forgetting~\cite{forgetting}. Details on the distribution of the collected post-training data are reported in App.~\ref{post_trained_data}.

We fine-tune the policy for five epochs using AdamW~\cite{adamw}, with a batch size of 24 and a learning rate of $10^{-5}$. We evaluate on three closed-loop benchmarks: Bench2Drive (B2D)~\cite{b2d}, CARLA Longest6 v2~\cite{carla_lb2}, and Fail2Drive (F2D)~\cite{f2d}. B2D evaluates in-distribution (ID) driving across diverse short routes, while Longest6 v2 stresses long-horizon closed-loop robustness with routes approximately $10\times$ longer and an additional penalty for driving substantially slower than local traffic. F2D 
finally evaluates out-of-distribution (OOD) generalization under unseen assets and novel scenario configurations. Following each benchmark's practices, we report Driving Score (DS) and Success Rate (SR) on B2D and F2D, and DS as well as Route Completion (RC) on Longest6 v2, where failure-free completion is substantially harder. Metrics are detailed in App.~\ref{details_metrics}. To evaluate takeover strategies across benchmarks
and analyze the effects of trigger design and expert
choice, we implement a TakeVLA-style baseline
(TakeVLA-noRL) as its official implementation is not publicly
available. We adopt its takeover triggers
and PDM-Lite expert within our post-training
pipeline, while retaining the same pretrained student
and fine-tuning recipe as RoG-DAgger.
To focus the comparison on takeover-based data collection, we omit TakeVLA's Scenario Dreaming reinforcement learning stage. This stage is potentially complementary to our approach and could also be applied after RoG-DAgger post-training, which we leave for future work. Further details are in App.~\ref{app:takevla_reproduction}.
We report the mean and standard deviation over
three independent training runs for RoG-DAgger
and TakeVLA-noRL.
\subsection{ID and Long-Horizon Benchmarks}\label{id_benchmark}
We report results on B2D and Longest6 v2 in Tab.~\ref{tab:b2d_longest6}. Despite its richer solution space, RoG-Expert is outperformed by PDM-Lite on both benchmarks. This is partly expected because RoG-Expert restricts privileged information to the student's field of view, sacrificing some standalone expert performance in favor of student-compatible supervision. Nevertheless, post-training substantially improves the base policy. RoG-DAgger achieves 90.11 DS and 73.29\% SR, outperforming the original SimLingo baseline by 5.0 DS points and 6.0 p.p.~in SR. Under the same VLM backbones, RoG-DAgger consistently improves over SimLingo (see App.~\ref{model_details}). This demonstrates that our post-training procedure effectively narrows the open-loop-closed-loop distribution gap. Compared to published post-training methods, RoG-DAgger clearly outperforms TakeAD and MindDrive and performs comparably to TakeVLA. Our TakeVLA-noRL baseline achieves lower performance than the published TakeVLA (87.61 vs.~89.72 DS), reflecting differences in the training frameworks, including the absence of reinforcement learning. Under the same post-training pipeline, RoG-DAgger outperforms TakeVLA-noRL by 2.5 DS points and 5.9 p.p. in SR.

On Longest6 v2, RoG-DAgger achieves 45 DS and 83 RC, compared with 22 DS and 70 RC for the original SimLingo baseline, substantially outperforming other non-privileged methods. Compared with TakeVLA-noRL, RoG-DAgger achieves a higher DS (45 vs.~36) at the same RC (83), suggesting fewer or less severe driving infractions.

In Fig.~\ref{qua_example_1}, we provide a qualitative comparison of RoG-DAgger with SimLingo on a B2D scene where RoG-DAgger decelerates and shifts left to safely avoid the cross-traffic, whereas SimLingo continues accelerating and ultimately collides. Further qualitative examples and more detailed B2D metrics can be found in App.~\ref{app:in_dis}.
\subsection{Out-of-Distribution Benchmarks}\label{ood_benchmark}
In addition to in-distribution evaluation, we assess OOD generalization on F2D. As shown in Tab.~\ref{tab:fail2drive}, RoG-DAgger slightly outperforms SimLingo on the base scenarios, while TF++ achieves the highest DS on the generalization scenarios, potentially benefiting from its LiDAR input.

On the generalization counterpart, RoG-DAgger outperforms SimLingo by about 2.5 in DS and 8 p.p.~in SR. Compared with TakeVLA-noRL, RoG-DAgger further achieves 4.7~p.p.~higher SR.
In challenging scenarios such as \textit{FullyBlocked}, RoG-DAgger slows down and stops when the road is obstructed, whereas SimLingo often maintains speed and collides with the blocking vehicle. Although this exact scenario is absent in the CARLA LB2, we hypothesize that by learning from preventive pre-takeover samples encountered under its own policy, the model acquires more general behaviors, such as slowing down or stopping when an obstacle blocks the drivable path, rather than memorizing responses to specific training scenarios. Additional qualitative examples and more detailed F2D metrics are provided in App.~\ref{app:ood_eval}.
\subsection{Ablation Studies}\label{para:ablation_studies}

\begin{table}[t]
    \centering
    \caption{\textbf{Ablation studies on Bench2Drive.}
    To reduce computational cost, ablation results are averaged over three evaluation runs using a single checkpoint per configuration.}
    \label{tab:b2d_ablation}
    \small
    \resizebox{1\columnwidth}{!}{
    \begin{tabular}{
        @{}l
        r@{}S[table-format=-2.2]
        r@{}S[table-format=-2.2]
        @{}
    }
        \toprule
        \textbf{Model}
        & \multicolumn{2}{c}{\textbf{DS} $\uparrow$}
        & \multicolumn{2}{c}{\textbf{SR} $\uparrow$} \\
        \midrule
        RoG-DAgger (full)
        & & 90.11
        & & 73.29 \\
        \midrule
        \quad $-$ Trajectory-and-speed expansion
        & $\Delta=$ & -2.72
        & $\Delta=$ & -5.61 \\
        \quad $-$ Solvability-aware collection
        & $\Delta=$ & -3.39
        & $\Delta=$ & -7.70 \\
        \quad $-$ FoV alignment
        & $\Delta=$ & -0.61
        & $\Delta=$ & -2.93 \\
        \midrule
        \quad $-$ Pre-takeover data
        & $\Delta=$ & -2.35
        & $\Delta=$ & -5.56 \\
        \bottomrule
    \end{tabular}}
\end{table}
We validate our design choices on B2D, with results summarized
in Tab.~\ref{tab:b2d_ablation}. First, we individually ablate
the three main components of RoG-DAgger. Removing any component
reduces both DS and SR relative to the full method. Disabling
solvability-aware takeover collection produces the largest performance
degradation (3.39 DS / 7.70 SR), highlighting the importance
of feasibility-aware intervention timing. Removing trajectory-and-speed
solution-space expansion also substantially reduces performance
(2.72 DS / 5.61 SR), suggesting the value of geometric adaptation
when speed adjustment alone is insufficient. Removing FoV
alignment yields smaller reductions (0.61 DS / 2.93 SR),
suggesting that aligning expert supervision with the student's
observable scene contributes to driving performance.
Next, we evaluate the contribution of preventive pre-takeover
supervision by training with post-takeover samples only.
Removing pre-takeover samples reduces B2D performance by
2.35 DS and 5.56 p.p.~in SR. This suggests that preventive supervision on policy-induced states preceding safety-critical events contributes substantially to closed-loop driving performance, beyond learning from post-takeover demonstrations alone. To control for training steps and data volume, we provide additional ablations in App.~\ref{app:add_ablation_studies}. These experiments help isolate the contributions of components from differences in training conditions.

\subsection{Takeover Trigger Timing Analysis}
\label{sec:trigger_timing}
To better understand the performance gap between RoG-DAgger and TakeVLA-noRL under the same post-training pipeline, we investigate the roles of expert choice and takeover timing. We first replace the PDM-Lite expert
in TakeVLA-noRL with RoG-Expert while retaining
its takeover triggers. The RoG-Expert variant achieves 86.75 $\pm$ 1.40 DS on B2D, compared with 87.61 $\pm$ 0.26 DS for the original PDM-Lite variant, showing no improvement from expert substitution alone. Notably, TakeVLA-noRL collects approximately three times as many samples as RoG-DAgger. We hypothesize that more frequent, premature takeovers alter the subsequent ego-state distribution, limiting exposure to safety-critical policy-induced states where improved expert supervision using RoG-Expert may be most beneficial.

To examine the trigger behavior underlying this hypothesis, we conduct a post-hoc analysis of takeover frequency and timing on identical student-policy rollouts, without executing expert takeovers.
We compare steering-discrepancy~\cite{takead}, insufficient student deceleration~\cite{takevla},
and constant-velocity TTC~\cite{takevla} triggers
against our solvability-aware trigger.
We record the first activation of each trigger
relative to the observed collision time, considering
the 2\,s pre-collision window as the target region
for preventive supervision.
As shown in Tab.~\ref{tab:trigger_timing}, on collision-ending episodes, our solvability-aware trigger achieves a temporal precision of 0.82 and an in-window recall of 0.65, compared with 0.51 and 0.50, respectively, for the strongest alternative, constant-velocity TTC.  Our trigger also produces \mbox{2--3$\times$} fewer
activations in collision-free episodes.
Moreover, its in-window activations are concentrated
closer to the observed collision
(Fig.~\ref{fig:trigger_comparison}).

These findings suggest that discrepancy-based and heuristic predictive triggers may select states where
intervention is not yet necessary. In particular, expert-student disagreement does not necessarily indicate imminent failure, given the multi-modality of valid driving behaviors, while constant-velocity TTC neglects potential braking and steering responses. 
In contrast, our rollout-guided trigger evaluates
nominal braking sufficiency and corrective-plan
feasibility, targeting policy-induced states where
preventive replanning becomes necessary.
Together with the expert-substitution experiment,
this analysis suggests that solvability-aware
trigger selection contributes to RoG-DAgger's
performance advantage. Additional details are
provided in App.~\ref{app:comparison_with_triggers}.

%% file: sec/5_discussion.tex
\section{Discussion}
In this work, we introduced RoG-DAgger, a post-training framework that improves pre-trained driving policies through closed-loop expert supervision. To mitigate the open-loop-closed-loop distribution shift, RoG-DAgger extends conventional DAgger with trajectory-and-speed solution-space expansion, solvability-aware triggering, and expert-student field-of-view alignment, while validating expert demonstrations through kinematic rollouts. Applied to SimLingo, RoG-DAgger consistently improves closed-loop performance on multiple benchmarks. These results demonstrate that learning from policy-induced states with reliable expert supervision can substantially improve closed-loop driving performance, particularly in challenging states that are underrepresented during pre-training, highlighting a promising direction for post-training of autonomous agents, e.g., for general robotics.

Nevertheless, RoG-DAgger's rollout-based solvability estimation remains approximate. Following prior work, our actor rollout assumes that surrounding actors remain non-reactive over a short time horizon, which can lead to inaccurate forecasts when they change their driving intentions. In addition, although RoG-DAgger expands the expert's trajectory-and-speed solution space, the considered candidate set 
may not contain all feasible collision-avoidance solutions. 
Extending RoG-DAgger with interaction-aware actor prediction, including robust uncertainty estimates~\cite{sicking2021novel, pintz2022survey} and richer trajectory generation is an important direction for future work.

A further constraint of our method is its reliance on CARLA closed-loop simulation, which limits direct application to real-world data. Extending rollout-guided post-training to high-fidelity data-driven simulators (e.g., AlpaSim~\cite{alpasim}) is a natural direction for future research.

%% file: sec/X_suppl.tex
\clearpage
\maketitlesupplementary
\appendix
\setlength{\textfloatsep}{8pt plus 2pt minus 2pt}
\setlength{\floatsep}{8pt plus 2pt minus 2pt}
\setlength{\intextsep}{8pt plus 2pt minus 2pt}
\section{Model Architecture, Pre-Training and Backbone Comparison}\label{model_details}
SimLingo~\cite{simlingo} operates on a front-facing camera view with a horizontal field of view (FoV) of $110^\circ$, which defines the visual context available to the policy for scene understanding and planning. For motion prediction, the policy outputs 10 trajectory waypoints together with 10 corresponding speed waypoints, jointly representing the spatial path and desired speed profile over the prediction horizon. These planning outputs are decoded from the learned visual-language representation using lightweight MLP heads for trajectory and speed prediction.

With the model architecture established, we next describe the training procedure and optimization settings used for pre-training. We pre-train the model following the SimLingo training recipe, while keeping all other unspecified settings identical to those in SimLingo. Training is performed for 5 epochs with a learning rate of $1\times10^{-4}$ and an effective batch size of 48 across 4 GPUs, corresponding to 12 samples per GPU. We apply LoRA~\cite{lora} with rank $r=32$ and scaling factor $\alpha=64$. Although SimLingo enables chain-of-thought~\cite{cot} output by default, the original paper reports no notable performance degradation when it is disabled~\cite{simlingo}. We observe the same behavior in our implementation and thus disable the chain-of-thought output in all experiments to reduce inference latency. 
Moreover, we replace the InternVL backbone used in \cite{simlingo} with a more recent Qwen3-VL-2B-Instruct.
As shown in Tab.~\ref{tab:simlingo_qwen_comparison},
the original InternVL-based SimLingo and our
Qwen3-VL-based implementation achieve comparable
B2D performance. Throughout the main benchmark
tables, SimLingo refers to the published results
of the original InternVL-based model, whereas
our main post-training experiments and qualitative
comparisons use the Qwen3-VL-based implementation
(SimLingo-Qwen).

To assess the effectiveness of RoG-DAgger
independently of the backbone replacement, we
additionally apply it to the original InternVL-based
SimLingo checkpoint (see Tab.~\ref{tab:simlingo_qwen_comparison}). RoG-DAgger improves B2D
performance by 4.14~DS points and 6.48~p.p.\ in SR
on InternVL, with corresponding gains of
4.13~DS points and 8.14~p.p.\ on Qwen3-VL.
These results indicate that the performance gains do not stem from the backbone replacement. This finding complements VLM4VLA~\cite{vlm4vla},
which shows that general-purpose VLM capabilities
do not necessarily translate into better
downstream control performance.

We analyze the B2D failure cases in Tab.~\ref{tab:failure_breakdown} and find that collisions account for roughly 80\% of the analyzed failures. We exclude the \textit{YieldToEmergencyVehicle} scenario from this analysis because SimLingo lacks a rear-facing camera and therefore cannot reliably detect and proactively yield to an emergency vehicle approaching \mbox{from behind}.
\setcounter{table}{3}

\begin{table}[h]
    \centering
    \caption{\textbf{Comparison between the original SimLingo
    and our Qwen3-VL-2B-based version on B2D.} DS and SR denote
    driving score and success rate, respectively. RoG-DAgger
    improves SimLingo under both VLM backbones.}
    \resizebox{\linewidth}{!}{
    \begin{tabular}{lcccc}
        \toprule
        \textbf{Model}
        & \textbf{DS}
        & \textbf{$\Delta$ DS}
        & \textbf{SR}
        & \makecell{\textbf{$\Delta$ SR}\\\textbf{(p.p.)}} \\
        \midrule
        SimLingo~\cite{simlingo}
        & $85.07\,{\scriptstyle\pm 0.95}$ & --
        & $67.27\,{\scriptstyle\pm 2.11}$ & -- \\

        + RoG-DAgger (\textbf{ours})
        & $89.21\,{\scriptstyle\pm 1.98}$ & $+4.14$
        & $73.75\,{\scriptstyle\pm 2.73}$ & $+6.48$ \\
        \midrule
        SimLingo-Qwen3
        & $85.98\,{\scriptstyle\pm 0.91}$ & --
        & $65.15\,{\scriptstyle\pm 2.05}$ & -- \\

        + RoG-DAgger (\textbf{ours})
        & $90.11\,{\scriptstyle\pm 0.20}$ & $+4.13$
        & $73.29\,{\scriptstyle\pm 1.38}$ & $+8.14$ \\
        \bottomrule
    \end{tabular}}
    \label{tab:simlingo_qwen_comparison}
\end{table}

\begin{table}[t]
\centering
\caption{\textbf{Breakdown of the recorded failures.} After excluding five
\textit{Yield To Emergency Vehicle} cases, we analyze 68 failures, of which
53 are collision-related. The collision subcategories are not mutually
exclusive, as one failure involves both a vehicle and a static object.}
\begin{tabular}{lrr}
\toprule
\textbf{Failure Category} &
\textbf{Count} &
\textbf{Proportion (\%)} \\
\midrule
Total recorded failures             & 73 & -- \\
Ignored cases     & 5  & -- \\
\textbf{Analyzed failures}          & \textbf{68} & \textbf{100.0} \\
\midrule
\textbf{Collision-related failures} & \textbf{53} & \textbf{77.9} \\
\quad Collision with vehicle        & 49 & 72.1 \\
\quad Collision with static         & 4  & 5.9 \\
\quad Collision with pedestrian     & 1  & 1.5 \\
\textbf{Non-collision failures}              & \textbf{15} & \textbf{22.1} \\
\bottomrule
\end{tabular}

\label{tab:failure_breakdown}
\end{table}

\section{Implementation Details of RoG-DAgger}\label{rog_expert_details}

This section provides additional experimental analyses
and implementation details of RoG-DAgger.
We first present a preliminary DAgger experiment in App.~\ref{pre_pdm_dagger}
and a subsequent expert solvability analysis
in App.~\ref{app:expert_solvability_analysis}, motivating
expert solution-space expansion and
solvability-aware takeover timing, respectively. Next we present the kinematic
rollout model and its TTC prediction accuracy
in App.~\ref{app:kinematic_ego_rollout}.
We then compare takeover triggers and evaluate
solvability-aware takeover timing in
Apps.~\ref{app:comparison_with_triggers}
and~\ref{app:solvability_aware_timing}.
Next, we detail the data-collection trigger conditions
and trajectory-and-speed solution-space expansion
in Apps.~\ref{trigger_condition}
and~\ref{traj_and_speed_expansion}, respectively.
We further provide qualitative examples in
Apps.~\ref{para:examples_geo_shifting}
and~\ref{para:solvability_aware_examples}.
Finally, we describe rule-based data filtering
and additional implementation details in
Apps.~\ref{data_filtering}
and~\ref{para:further_details}.

\subsection{Preliminary DAgger Experiment}\label{pre_pdm_dagger}
As described in Sec.~\ref{rog_expert}, we conduct
a preliminary DAgger run on CARLA LB2 training
routes using PDM-Lite as the expert and observed
collisions as takeover triggers. To examine
whether PDM-Lite can provide feasible supervision
at failure-proximal policy-induced states, we
retrospectively query the expert at each
pre-takeover state and evaluate its predicted
trajectory-and-speed plan using our
controller-in-the-loop kinematic rollout model.
A plan is considered rollout-solvable if its
predicted ego trajectory is collision-free
with respect to surrounding actors.

We consider pre-collision windows of 20 frames,
following TakeVLA~\cite{takevla}, and 40 frames,
corresponding to our default setting.
As shown in the first two rows of
Tab.~\ref{tab:pre_post_collision_stats},
PDM-Lite achieves rollout-solvable rates of
55.09\% and 71.55\%, respectively. Thus,
44.91\% and 28.45\% of its plans remain
collision-prone under the rollout model.
These findings suggest that the restricted
solution space of PDM-Lite can limit its
ability to provide feasible corrective
supervision at policy-induced states,
motivating our expert solution-space~expansion.

\begin{table}[t!]
\centering
\caption{\textbf{Rollout-solvable rates of expert
plans at pre-takeover states.}
PDM-Lite and RoG-Expert are evaluated on identical
policy-induced states from the 20- and 40-frame
pre-collision windows. A plan is rollout-solvable
if no collision is predicted under the same
kinematic rollout model.}
\label{tab:pre_post_collision_stats}

\resizebox{\columnwidth}{!}{%
\begin{tabular}{@{}lrrrr@{}}
\toprule
\multirow{2}{*}{\textbf{Expert}} &
\multirow{2}{*}{\textbf{Window}} &
\multicolumn{1}{c}{\textbf{Total}} &
\multicolumn{1}{c}{\textbf{Solvable}} &
\multirow{2}{*}{\textbf{Rate}} \\
& &
\multicolumn{1}{c}{\textbf{Samples}} &
\multicolumn{1}{c}{\textbf{Samples}} &
\\
\midrule

PDM-Lite
& 20 & 7,226 & 3,981 & 55.09\% \\

PDM-Lite
& 40 & 14,397 & 10,301 & 71.55\% \\

\midrule

RoG-Expert
& 20 & 7,226 & 5,021 & 69.49\% \\

RoG-Expert
& 40 & 14,397 & 11,728 & 81.46\% \\

\bottomrule
\end{tabular}%
}
\end{table}

\subsection{Expert Solvability Analysis}\label{app:expert_solvability_analysis}
To assess whether RoG-Expert addresses this
limitation, we retrospectively query it at
the same pre-takeover states in App.~\ref{pre_pdm_dagger} and evaluate
its plans using the same kinematic rollout
model. As shown in the last two rows of
Tab.~\ref{tab:pre_post_collision_stats},
RoG-Expert increases the rollout-solvable
rates to 69.49\% and 81.46\% for the
20- and 40-frame windows, respectively,
improving upon PDM-Lite by 14.40 and
9.91 percentage points. These results
demonstrate the higher rollout solvability
of the RoG-Expert configuration at
identical policy-induced states.

\begin{figure}[t!]
\centering
\includegraphics[width=\linewidth]{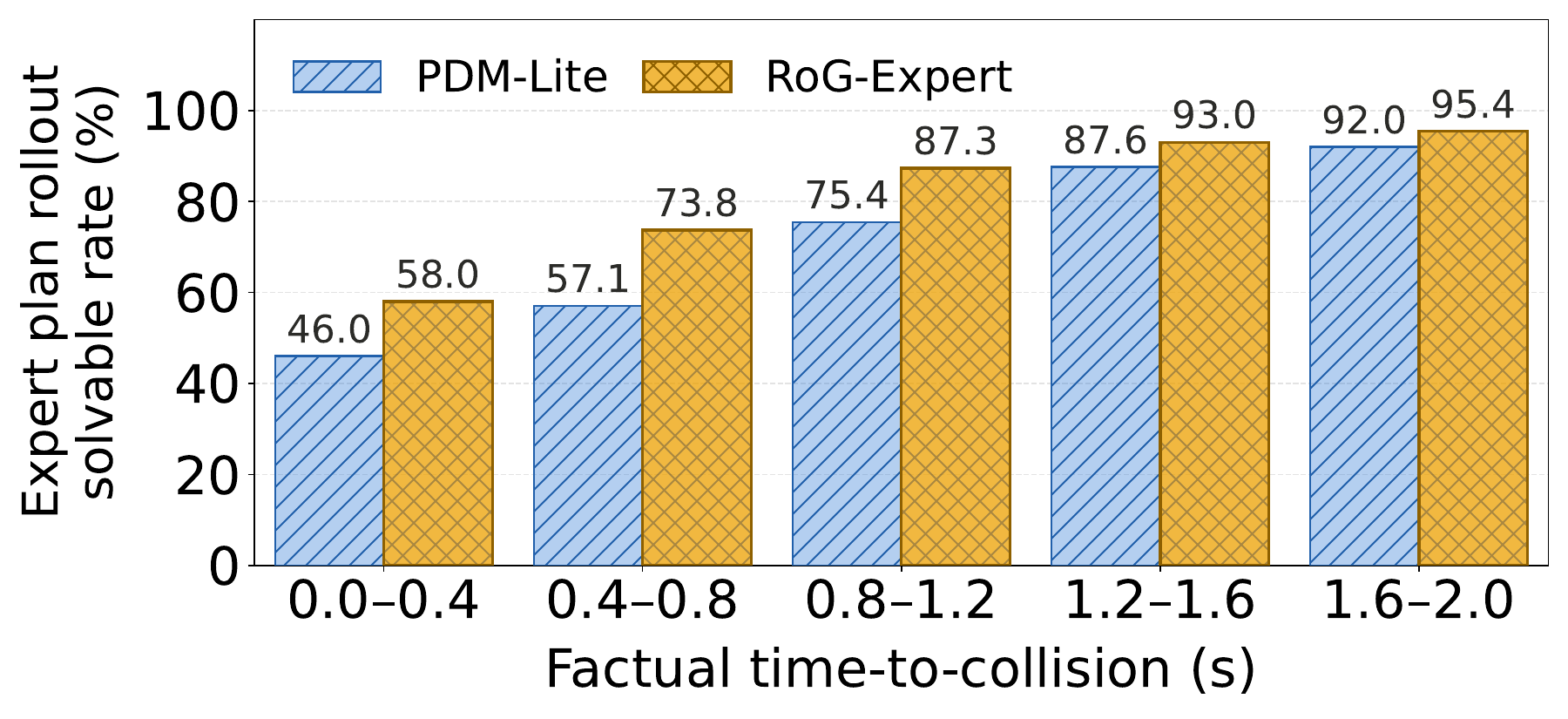}
\caption{\textbf{Expert plan rollout solvability
versus factual time-to-collision (TTC).}
Both experts are evaluated on identical
pre-takeover states, grouped by time remaining
until the observed collision. RoG-Expert achieves higher rollout-solvable rates than PDM-Lite across all TTC intervals, with larger improvements at lower TTC values. However, both experts exhibit decreasing solvability as the collision approaches.}
\label{fig:solvable_rate_vs_factual_ttc}
\end{figure}

Nevertheless, 30.51\% and 18.54\% of
RoG-Expert's plans remain collision-prone
in the respective windows. We further
analyze rollout solvability as a function
of factual time-to-collision (TTC).
As shown in
Fig.~\ref{fig:solvable_rate_vs_factual_ttc},
RoG-Expert consistently outperforms
PDM-Lite, with a larger advantage at
shorter TTC intervals. However, both
experts exhibit decreasing solvability
as the observed collision approaches.
For RoG-Expert, the rollout-solvable
rate drops from 95.4\% at 1.6--2.0\,s
before collision to 58.0\% in the
final 0.4\,s.

These findings suggest that, despite
improved expert planning, the opportunity
for feasible corrective supervision
diminishes near impending collisions,
motivating solvability-aware takeover
timing. We note that rollout solvability
is assessed using kinematic predictions
with non-reactive surrounding actors,
rather than observed collision outcomes
in CARLA.

\subsection{Kinematic Ego Rollout Model} \label{app:kinematic_ego_rollout}
We provide the pseudocode for our kinematic ego rollout model in Alg.~\ref{alg:ego_rollout}. Following PDM-Lite~\cite{pdm_lite}, we adopt a planning horizon of 2\,s in most cases and 1.1\,s during lane change. To assess its prediction accuracy, we conduct a diagnostic evaluation on CARLA LB2. Specifically, we roll out the ego vehicle using the student's predicted geometric trajectory and target speed, thereby accounting for both lateral and longitudinal motion from the current state. While our solvability-aware trigger evaluates candidate expert actions, this diagnostic experiment uses \textit{student}-predicted actions to assess the accuracy of the underlying rollout model. Bounding-box inflation is disabled during this evaluation (see App.~\ref{para:further_details}). 

\begin{figure}[t]
  \centering
  \includegraphics[width=0.8\columnwidth]{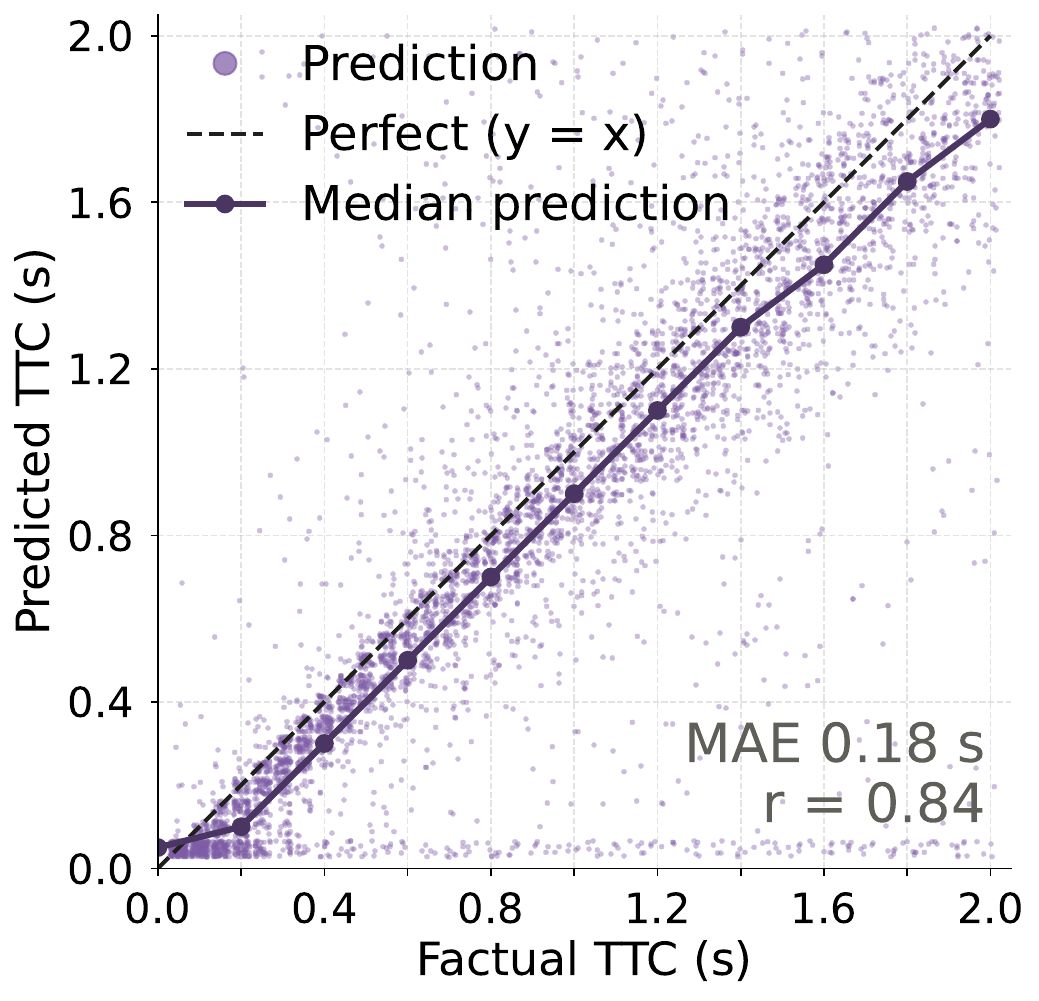}
  \caption{\textbf{TTC Prediction Accuracy of the Kinematic Ego Rollout Model.}
    Each point represents a simulation step within the final
    \SI{2}{\second} preceding an observed collision
    (5,362 samples). The rollout model estimates TTC
    using the student's predicted geometric trajectory
    and target speed, without bounding-box inflation.
    Predicted TTC is compared against factual TTC,
    defined as the actual time remaining until collision.
    The solid line shows the median predicted TTC in
    \SI{0.2}{\second} bins of factual TTC.
    The model achieves an MAE of \SI{0.18}{\second}
    and a Pearson correlation of $r=0.84$,
    with a slight tendency to underestimate TTC. This indicates a slightly more conservative risk estimation.}
  \label{fig:ttc_pred_vs_actual}
\end{figure}

A collision is predicted when the simulated ego vehicle's bounding box overlaps with that of another actor. For each simulation
step within 2\,s before an observed collision,
we estimate TTC using the applicable rollout
horizon. We evaluate steps where a collision
is predicted within this horizon, comparing the
predicted TTC with the factual TTC, defined as
the actual time remaining until collision. As shown in Fig.~\ref{fig:ttc_pred_vs_actual}, the resulting 5,362 samples yield a mean absolute error (MAE) of 0.18 s and a Pearson correlation coefficient of 0.84 between predicted and factual TTC, indicating a strong positive correlation and relatively low prediction error. The median predictions, computed in 0.2 s intervals, indicate a slight tendency to underestimate TTC, resulting in conservative estimates of collision risk. This discrepancy may partly arise from differences between our bounding-box-based collision criterion and CARLA's collision detection based on 3D mesh contact, although errors in the kinematic approximation may also contribute.
\begin{algorithm}[t]
\caption{\textbf{Kinematic Ego Rollout}}
\label{alg:ego_rollout} 

\textbf{Inputs}: ego state $\mathbf{s}_0^{\mathrm{ego}}$, geometric trajectory
$\mathbf{a}^{\mathrm{geo}}$, target speed $v_{\mathrm{target}}$, planning horizon $H$, braking ratio $\rho$\\
\textbf{Output}: ego rollout $\mathbf{a}_{\mathrm{rollout}}$

\begin{algorithmic}[1]

\STATE $\mathbf{s}_0 \leftarrow \mathbf{s}_0^{\mathrm{ego}},\;
\mathbf{a}_{\mathrm{rollout}} \leftarrow \emptyset$

\FOR{$k=0$ \TO $H-1$}
    \STATE $\phi_k \leftarrow
    \textsc{LateralCtrl}(\mathbf{s}_k,\mathbf{a}^{\mathrm{geo}})$

    \IF{$v_{\mathrm{target}} \approx 0
        ~\OR ~v_k > \rho\ v_{\mathrm{target}}$}
        \STATE $(u_k^{\mathrm{thr}},u_k^{\mathrm{brk}})
        \leftarrow (0,1)$
    \ELSE
        \STATE $(u_k^{\mathrm{thr}},u_k^{\mathrm{brk}})
        \leftarrow
        (\textsc{ThrottleCtrl}(v_k,v_{\mathrm{target}}),0)$
    \ENDIF

    \STATE $\mathbf{s}_{k+1}, v_{k+1} \leftarrow
    \textsc{BicycleModel}
    (\mathbf{s}_k,\phi_k,u_k^{\mathrm{thr}},u_k^{\mathrm{brk}})$

    \STATE $\mathbf{a}_{\mathrm{rollout}}[k+1] \leftarrow \mathbf{s}_{k+1}$
\ENDFOR

\STATE \textbf{return} $\mathbf{a}_{\mathrm{rollout}}$

\end{algorithmic}
\end{algorithm}

\vspace{5pt}

\subsection{Comparing Solvability-Aware Trigger with Other Triggers}\label{app:comparison_with_triggers}
As discussed in Sec.~\ref{sec:trigger_timing},
we investigate takeover frequency and timing
through a post-hoc analysis of student-policy
trajectories recorded during infraction-based
data collection on CARLA LB2.
The compared triggers are evaluated without
executing their takeovers; their activations
are only recorded. Expert takeover is performed
after observed collisions to recover and
continue the route. Thus, all compared triggers
are evaluated on identical student-policy
trajectories without altering their behavior
before collision.
We compare steering-discrepancy~\cite{takead},
insufficient student deceleration~\cite{takevla},
constant-velocity TTC~\cite{takevla}, and our
solvability-aware trigger. The evaluated triggers
do not initiate expert takeover.

An episode begins at route initialization or
when student control resumes after expert
recovery from a previous collision.
It ends at the next collision (collision-ending
episode) or when the route terminates without
a collision (collision-free episode).
For collision-ending episodes, we record the first
activation of each trigger relative to the observed
collision time, using the final 2\,s before collision
as the target window for preventive takeover.

We exclude 58 episodes involving static obstacles
not represented by the expert planner (e.g.,
vegetation) or dynamic actors that remain outside field of view during the 2\,s
pre-collision window. This leaves 276 collision-ending
episodes and 699 collision-free episodes.
As shown in Tab.~\ref{tab:trigger_timing},
our solvability-aware trigger achieves higher
temporal precision (0.82) and in-window recall
(0.65) than the strongest alternative,
constant-velocity TTC (precision:\ 0.51, recall:\ 0.50).
On collision-free episodes, our trigger produces activations in 204 episodes, compared with 416--660 for the
alternative triggers, a reduction of 51--69\%.

The solvability-aware trigger does not activate
in 56 collision-ending episodes. These misses
primarily involve lateral contacts with roadside
obstacles (e.g., opened vehicle doors extending
beyond CARLA's vehicle bounding boxes) and
conflicts with lead vehicles handled by the
expert's car-following logic rather than its
collision-avoidance module. These cases reveal
limitations in the expert's collision representation
and prediction. Nevertheless, our complementary
infraction-based collection protocol captures
observed failures for retrospective supervision,
providing coverage of failure types missed by
the anticipatory trigger.

\begin{figure}[t]
\centering
\includegraphics[width=1.0\columnwidth]{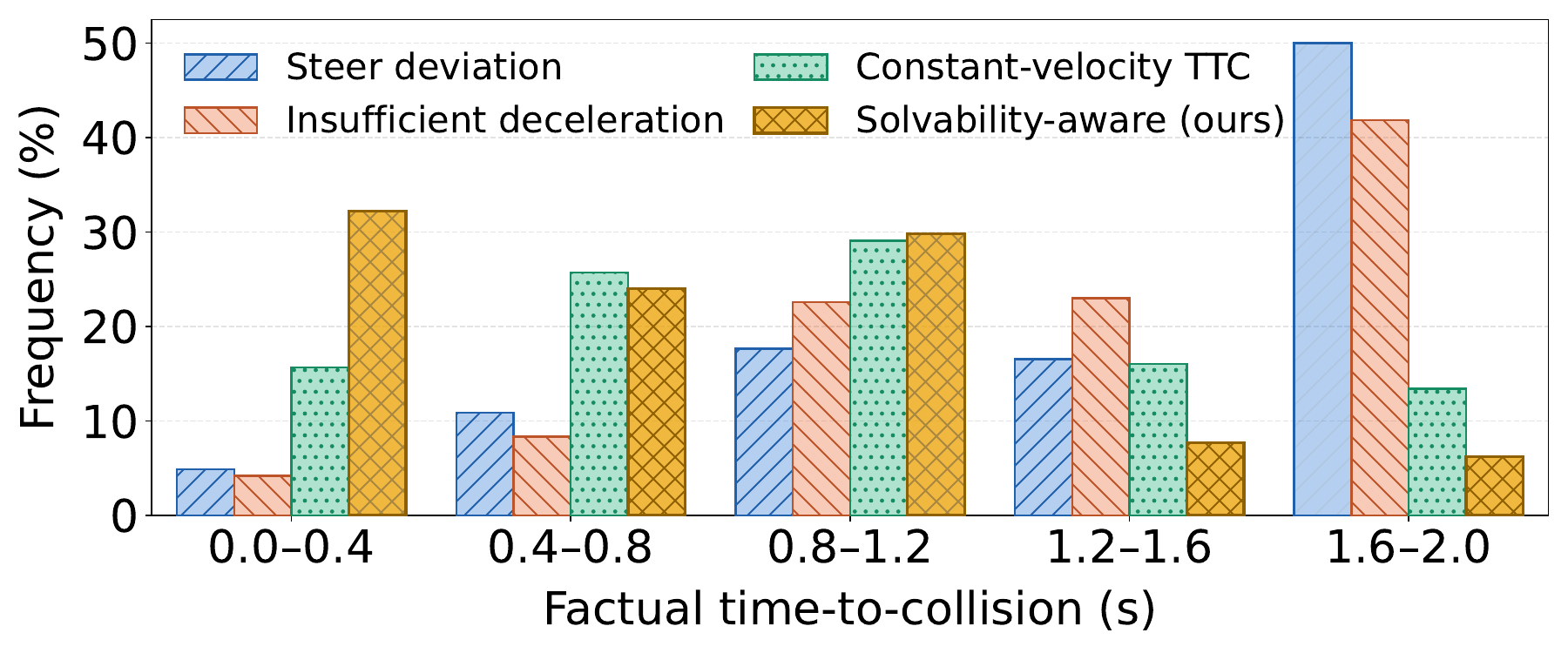}
\caption{\textbf{Distribution of the factual time-to-collision at each
trigger's first activation within the final 2\,s before collision,
normalized per trigger.} Our solvability-aware trigger activates most
often in the last 0.4\,s (32\%) and rarely at the start of the window
(6\% in 1.6-2.0\,s). In contrast, steering-deviation and
insufficient-deceleration triggers activate mostly at the start
of the window (50\% and 42\%).}
\label{fig:trigger_comparison}
\end{figure}

We further examine the timing of true-positive
activations within the 2\,s pre-collision window. 
The median factual
time-to-collision at first activation is 0.72\,s for our
solvability-aware trigger, compared with 0.95\,s for
constant-velocity TTC, 1.45\,s for insufficient deceleration,
and 1.62\,s for steering discrepancy.
As shown in Fig.~\ref{fig:trigger_comparison}, 
our solvability-aware trigger tends to activate
closer to the observed collision than the
alternative triggers.

These findings suggest that discrepancy-based
and heuristic predictive triggers may activate
before preventive intervention becomes necessary.
Expert-student disagreement does not necessarily
indicate imminent failure, given the multi-modality
of valid driving behaviors, while constant-velocity
TTC neglects potential braking and steering
responses. In contrast, our rollout-guided trigger
assesses nominal braking sufficiency and
corrective-plan feasibility, targeting
safety-critical states where preventive
intervention becomes necessary.
\begin{table}[t]
  \centering
  \caption{\textbf{Takeover-Trigger Timing on Collision Episodes.} A~trigger's first
  activation counts as a true positive (TP) if it occurs within 2\,s before the collision and as
  a false positive (FP) if it occurs earlier; FN denotes episodes in which the trigger never
  activates. Recall is $\mathrm{TP}/N$ and precision is $\mathrm{TP}/(\mathrm{TP}{+}\mathrm{FP})$.
  The last column counts activations in episodes without a collision.}
  \label{tab:trigger_timing}
  \resizebox{\columnwidth}{!}{%
  \begin{tabular}{lrrrrcc}
    \toprule
    \textbf{Trigger}
    & \textbf{TP}\,$\uparrow$
    & \textbf{FP}\,$\downarrow$
    & \textbf{FN}\,$\downarrow$
    & \textbf{Prec.}\,$\uparrow$
    & \textbf{Recall}\,$\uparrow$
    & \textbf{No-coll.}\,$\downarrow$ \\
    \midrule
    Steer deviation
    & 39 & 235 & 2 & 0.14 & 0.14 & 618 \\
    Insufficient deceleration
    & 14 & 259 & 3 & 0.05 & 0.05 & 660 \\
    Constant-velocity TTC
    & 138 & 133 & 5 & 0.51 & 0.50 & 416 \\
    Solvability-aware (\textbf{ours})
    & \textbf{180} & \textbf{40} & 56
    & \textbf{0.82} & \textbf{0.65} & \textbf{204} \\
    \bottomrule
  \end{tabular}%
  }
\end{table}

\subsection{Effectiveness of Solvability-Aware Trigger}\label{app:solvability_aware_timing}
As discussed in App.~\ref{app:comparison_with_triggers},
our solvability-aware trigger selects safety-critical
states while reducing premature activations.
Here, we further assess whether the proposed takeover
timing preserves sufficient opportunity for successful
expert intervention. We conduct experiments on the
same CARLA LB2 training routes, varying the takeover
timing of \texttt{Replan Success} and
\texttt{Repeated Replan Success} events, for which
the rollout identifies a feasible corrective plan.
We consider offsets of $\pm 0.15$ and $\pm 0.30$\,s.
For positive offsets, the student remains in
control during the imposed delay. Any collision
before the delayed takeover is counted as a
failure. On the contrary, negative offsets increase the TTC
thresholds to allow earlier takeover. For events reaching expert takeover,
we monitor collisions over the subsequent 2\,s.

\begin{figure}[t]
  \centering
  \includegraphics[width=\columnwidth]
  {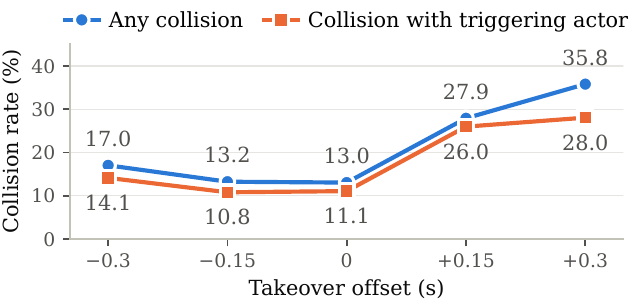}
  \caption{
  \textbf{Effect of takeover timing on observed collision rates.}
  We vary the activation timing of successful-replan
  triggers by $\pm 0.15$ and $\pm 0.30$\,s.
  Positive offsets delay expert takeover, whereas
  negative offsets increase the TTC thresholds to
  permit earlier intervention.
  Zero denotes the default timing. Collisions during a delayed takeover are counted
as failures; otherwise, collision outcomes are
evaluated over the 2\,s following expert takeover. We consider either
  any collision or only collisions with the
  triggering actor. Repeated triggers involving
  the same actor are counted once.
  Delaying takeover by 0.3\,s increases the
  triggering-actor collision rate from
  11.1\% to 28.0\%.
  }
  \label{fig:takeover_offset_collision_rate}
\end{figure}

As shown in Fig.~\ref{fig:takeover_offset_collision_rate},
delaying takeover by 0.3\,s increases the collision
rate with the triggering actor from 11.1\% to 28.0\%,
whereas earlier intervention provides no clear
improvement over the default timing. These results
suggest that the proposed trigger identifies a
favorable intervention window: delaying takeover
substantially reduces the opportunity for successful
correction, while earlier intervention offers limited
benefit. The nonzero collision rate at the default
timing reflects limitations of our rollout-based
prediction, potentially including unmodeled vehicle
motion during sharp turns (e.g., body sway) and static obstacles
excluded from expert planning.

We additionally evaluate \texttt{Replan Failure}
events over these runs. The observed collision rates
are $17.1 \pm 2.7$\% with the triggering actor
and $21.4 \pm 2.2$\% with any object.
Although higher than those of successful-replan
events, most replan-failed cases do not result in
observed collisions. This is consistent with the
conservative collision checking used by our
rollout model, which adopts inflated bounding
boxes following PDM-Lite. Thus, failure to find
a rollout-feasible plan does not necessarily imply
an unavoidable collision. Samples involving
observed collisions in the solvability-aware
collection run are excluded from training.

\subsection{Trigger Conditions}\label{trigger_condition}
\begin{figure*}[t]
    \centering
    \includegraphics[width=\textwidth]{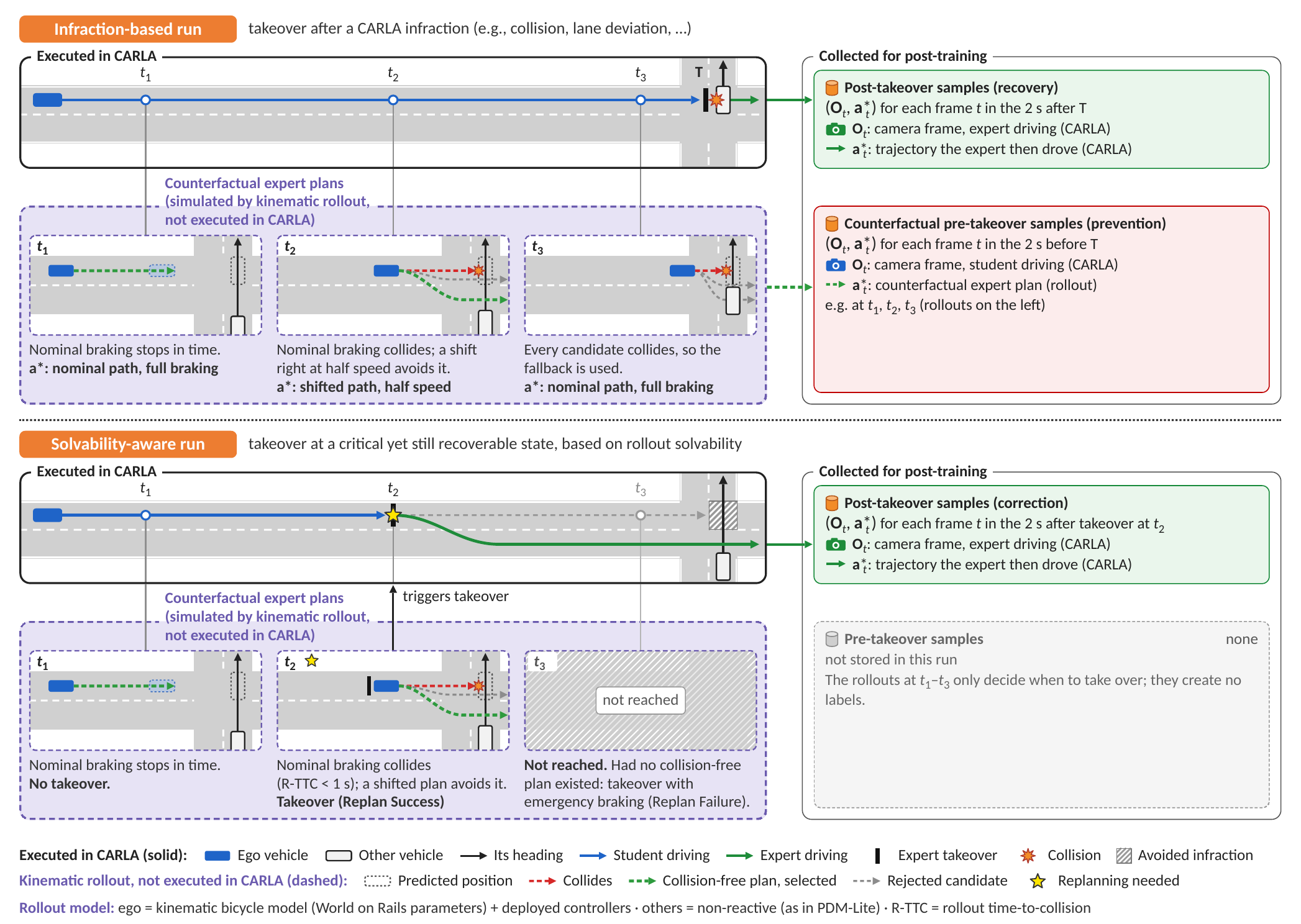}
    \caption{\textbf{Overview of the RoG-DAgger data-collection pipeline.}
    We conduct two independent data-collection runs.
    Top: Infraction-based collection.
    The expert takes over following a CARLA infraction.
    We collect post-takeover recovery demonstrations from
    expert-controlled driving and counterfactual pre-takeover
    supervision by querying the expert at preceding student states.
    The corresponding expert plans are evaluated using
    kinematic rollouts without execution in CARLA.
    Bottom: Solvability-aware collection.
    Counterfactual expert rollouts are used to identify
    critical yet recoverable states and trigger timely intervention.
    Only post-takeover expert demonstrations are retained
    for training in this run.
    Solid trajectories represent execution in CARLA,
    whereas dashed trajectories indicate simulated
    counterfactual rollouts.
    For clarity, the repeated replan success trigger
    and restart trigger are omitted from the illustration.}
    \label{fig:rog_dagger_data_collection}
\end{figure*}

As described in Sec.~\ref{rog_dagger_framework}, we conduct two independent data-collection runs using three types of triggers. The corresponding trigger conditions are detailed in Tab.~\ref{tab:collection_trigger_conditions}. For infraction-triggered events, we retain a 40-frame (2\,s) pre-takeover window and a 40-frame (2\,s) post-takeover window. For solvability-aware and restart triggers, we retain only post-takeover observations: 40 frames (2\,s) for solvability-aware interventions and 80 frames (4\,s) for restart events to capture the full recovery process. 
As illustrated in Fig.~\ref{fig:rog_dagger_data_collection},
RoG-DAgger employs two complementary data-collection
runs to obtain preventive and \mbox{corrective supervision.}
\begin{table}[t]
    \centering
    \caption{
       \textbf{ Details on the three types of trigger conditions used for post-training data collection of RoG-DAgger.}
        R-TTC denotes the predicted time-to-collision under nominal emergency braking, estimated by our 
        rollout model before \mbox{corrective replanning.}
    }
    \begin{tabular} {p{0.2\columnwidth} p{0.7\columnwidth}}
        \toprule
        \textbf{Trigger} & \textbf{Trigger Condition} \\
        \midrule

        \textbf{Infraction}
        &
        A standard CARLA infraction, including a collision, red-light violation, stop-sign violation
        or out-of-lane violation.
        \\

        \midrule

        \textbf{Solvability-Aware Takeover}
        &
                \texttt{Replan Success:} \newline
        $\mathrm{R\text{-}TTC} < 1\,\mathrm{s}$ $\mathtt{AND}$
        replanning succeeds.
        \par\vspace{0.3em}
        \texttt{Repeated Replan Success:}\newline
        $1\,\mathrm{s} < \mathrm{R\text{-}TTC} < 1.5\,\mathrm{s}$ $\mathtt{AND}$ \newline
        replanning succeeds in two of four consecutive frames.
        \par\vspace{0.3em}
        \texttt{Replan Failure:}\newline
        $\mathrm{R\text{-}TTC} < 1\,\mathrm{s}$ $\mathtt{AND}$ replanning fails.

        \\

        \midrule

        \textbf{Restart}
        &
        A persistent discrepancy is detected between the ego vehicle and the
        expert model: the ego speed remains below $0.1\,\mathrm{m/s}$ for all
        of the previous 200 frames, while both the mean expert target speed
        over these frames and the current expert target speed exceed
        $0.4\,\mathrm{m/s}$.
        \\

        \bottomrule
    \end{tabular}

    \label{tab:collection_trigger_conditions}
\end{table}
\subsection{Expansion~of~Trajectory-Speed \mbox{Solution Space}}\label{traj_and_speed_expansion}
\begin{algorithm}[t!]
\caption{\textbf{Trajectory-and-speed solution-space expansion.}
$\textsc{PDM-Collision}$ denotes PDM-Lite's native collision prediction used to
trigger emergency braking. It returns the collision-relevant actor $o$ and the
index set $\mathcal I$ of nominal-trajectory waypoints involved in the predicted
ego-actor overlap. $\textsc{Local-Shift}(\mathbf{a},\mathcal I,o,d)$ shifts
only the waypoints of ego trajectory $\mathbf{a}$ indexed by $\mathcal I$ by distance $d$
along the direction perpendicular to the local trajectory, choosing the side
away from actor $o$'s direction of motion, with smooth ease-in/ease-out transitions
to the unchanged trajectory. $\mathcal R_{\mathrm{ego}}$ denotes the
controller-in-the-loop ego rollout model (see Sec.~\ref{kinematic_rollout}). Replanning is targeted to avoid $o$, while candidate plans are validated against all predicted actors
$\mathbf{A}_t$. For brevity, Alg.~\ref{alg:solution_space} omits the planning status, which is tracked separately to distinguish successful nominal braking, successful replanning, and replanning failure.}
\label{alg:solution_space}

\textbf{Inputs}: ego state $\mathbf{s}_t^{\mathrm{ego}}$, nominal trajectory
$\mathbf{a}_t^{\mathrm{geo}}$, current \mbox{speed $v_t$}, predicted actors
$\mathbf{A}_t$, shift unit \mbox{size $\Delta d$}, maximum shift $d_{\max}$\\
\textbf{Output}: expert plan
$(\mathbf{a}_{\mathrm{geo}}^*,\mathbf{a}_{\mathrm{temp}}^*)$
or $\varnothing$

\begin{algorithmic}[1]

\STATE $(o,\mathcal I) \leftarrow
\textsc{PDM-Collision}
(\mathbf{s}_t^{\mathrm{ego}},
 \mathbf{a}_t^{\mathrm{geo}},
 \mathbf{A}_t)$

\IF{$o=\varnothing$}
    \STATE \textbf{return} $\varnothing$ 
\ENDIF

\STATE $\mathbf{a}_{\mathrm{temp}}^{\mathrm{brk}}
\leftarrow
\mathcal R_{\mathrm{ego}}
(\mathbf{s}_t^{\mathrm{ego}},
 \mathbf{a}_t^{\mathrm{geo}},0)$

\IF{$\textsc{Collision-Free}
(\mathbf{a}_{\mathrm{temp}}^{\mathrm{brk}},\mathbf{A}_t)$}
    \STATE \textbf{return}
    $(\mathbf{a}_t^{\mathrm{geo}},
      \mathbf{a}_{\mathrm{temp}}^{\mathrm{brk}})$
\ENDIF

\IF{$\mathbf{s}_t^{\mathrm{ego}}$ is inside a junction}
    \STATE $\mathcal V_{\mathrm{target}}
    \leftarrow \{0.5\,v_t,\,0.25\,v_t,\,0\}$
\ELSE
    \STATE $\mathcal V_{\mathrm{target}}
    \leftarrow \{0\}$
\ENDIF

\FOR{$v_{\mathrm{target}} \in \mathcal V_{\mathrm{target}}$}
    \FOR{$d=\Delta d,\,2\Delta d,\ldots,d_{\max}$}

        \STATE $\mathbf{a}_{\mathrm{geo}}'
        \leftarrow
        \textsc{Local-Shift}
        (\mathbf{a}_t^{\mathrm{geo}},\mathcal I,o,d)$

        \IF{$\textsc{Lane-Compliant}(\mathbf{a}_{\mathrm{geo}}')$}

            \STATE $\mathbf{a}_{\mathrm{temp}}'
            \leftarrow
            \mathcal R_{\mathrm{ego}}
            (\mathbf{s}_t^{\mathrm{ego}},
             \mathbf{a}_{\mathrm{geo}}',
             v_{\mathrm{target}})$

            \IF{$\textsc{Collision-Free}
            (\mathbf{a}_{\mathrm{temp}}',\mathbf{A}_t)$}
                \STATE \textbf{return}
                $(\mathbf{a}_{\mathrm{geo}}',
                  \mathbf{a}_{\mathrm{temp}}')$
            \ENDIF

        \ENDIF

    \ENDFOR
\ENDFOR

\STATE \textbf{return}
$(\mathbf{a}_t^{\mathrm{geo}},
  \mathbf{a}_{\mathrm{temp}}^{\mathrm{brk}})$

\end{algorithmic}
\end{algorithm}

When full braking alone is insufficient, we expand the solution space by generating geometric candidates around the predicted ego-actor overlap, as summarized in Alg.~\ref{alg:solution_space}. Replanning is triggered when the nominal PDM-Lite collision detector requests an emergency stop, which is the case if the ego vehicle is moving faster than $0.4 \mathrm{m/s}$, while the target speed falls below $10^{-3}\mathrm{m/s}$. Collision prediction is then performed assuming a target speed of zero.

Given the resulting predicted collision interval, we locally modify the affected portion of the nominal trajectory rather than shifting the entire route. In particular, the modified segment starts approximately in a distance of 
\begin{equation}
5\,\mathrm{m}+l_{\mathrm{half\_ego}}+l_{\mathrm{half\_actor}}
\end{equation}
before the first predicted ego-actor overlap and extends \mbox{beyond} the final predicted overlap by approximately 
\begin{equation}
2\,\mathrm{m}+l_{\mathrm{half\_ego}}+l_{\mathrm{half\_actor}\ ,}
\end{equation}
where $l_{\mathrm{half\_ego}}$ and $l_{\mathrm{half\_actor}}$ denote the longitudinal half-extents of the ego vehicle and collision-relevant actor, respectively. The $5\,\mathrm{m}$ pre-collision margin lets the avoidance maneuver begin before entering the predicted conflict region, while the $2\,\mathrm{m}$ post-collision extension avoids abrupt curvature and steering changes.

Within this segment, we shift the nominal trajectory laterally, i.e., perpendicular to the ego-route direction and away from the collision-relevant actor. To avoid abrupt geometric changes, the lateral displacement is introduced and removed using a spatial cosine ease-in/ease-out transition over 2 $\mathrm{m}$. Specifically, we define the transition weight as

\begin{equation}
E(u)=\frac{1-\cos(\pi u)}{2}, \qquad u\in[0,1],
\end{equation}

where $u$ denotes the normalized progress through the transition interval.
During the ease-in transition, the weight increases smoothly from $E(0)=0$ to
$E(1)=1$, and \mbox{the interpolated trajectory} is given by
\begin{equation}
\bigl(1-E(u)\bigr)~\mathbf{a}_{\mathrm{geo}}
+
E(u)~\mathbf{a}'_{\mathrm{geo}}\ ,
\end{equation}

where $\mathbf{a}'_\mathrm{geo}$ denotes the shifted trajectory.
For the ease-out transition, the same weighting profile is applied in reverse,
smoothly returning the shifted trajectory to the nominal trajectory. As a result, the trajectory gradually approaches the selected lateral displacement, remains shifted through the central collision-avoidance region, and smoothly merges back into the nominal trajectory after the predicted conflict.

As in Alg.~\ref{alg:solution_space}, we search over increasing lateral offsets and retain the first candidate whose rollout is both collision-free and lane-compliant. In the implementation, offsets $\Delta d$ are evaluated in $0.2~\mathrm{m}$ increments until $d_{\max} = 3~m$, thereby favoring the smallest geometric deviation that resolves the predicted collision.  A candidate is considered lane-valid only if its lateral deviation remains within approximately

\begin{equation}
\frac{w_{\mathrm{lane}}}{2}+0.2\,\mathrm{m},
\end{equation}

where $w_{\mathrm{lane}}$ is the local lane width. Candidates outside this admissible region are discarded. We first consider target-speed options in descending order and, for each target speed, evaluate geometric candidates with progressively increasing lateral offsets using the controller-in-the-loop rollout model $\mathcal R_{\mathrm{ego}}$, following Alg.~\ref{alg:ego_rollout}. Candidates are retained only when their temporal rollout remains collision-free with respect to all predicted surrounding actors. We prioritize higher target speeds and then smaller lateral deviations. If no valid shifted candidate exists, we fall back to the nominal geometric trajectory with full braking.

\subsection{Examples of Geometric Trajectory Shifting}\label{para:examples_geo_shifting}
In the main paper, we discuss why a restricted trajectory-and-speed solution space can lead an expert model to produce suboptimal actions. Here, we provide a concrete example in Figs.~\ref{fig:no_replan} and \ref{fig:replan_expert}. We evaluate two privileged experts in the \textit{MergerIntoSlowTraffic} scenario. The expert in Fig.~\ref{fig:no_replan} follows the default PDM-Lite strategy: it tracks a precomputed geometric trajectory and attempts to avoid collisions solely through full braking, i.e., by setting the target speed to zero. In contrast, the RoG-Expert in Fig.~\ref{fig:replan_expert} uses an expanded trajectory-and-speed solution space. While the default expert fails to complete the scenario safely, the RoG-Expert with the expanded solution space resolves the traffic conflict by combining adapted geometric trajectory and full braking. This comparison shows that an expert with a restricted trajectory-and-speed solution space can itself become the performance bottleneck, as it may fail to provide a safe corrective action even with privileged \mbox{scene information}.
\setcounter{figure}{9}
\begin{figure*}[p]
\centering
\includegraphics[width=2.0\columnwidth]{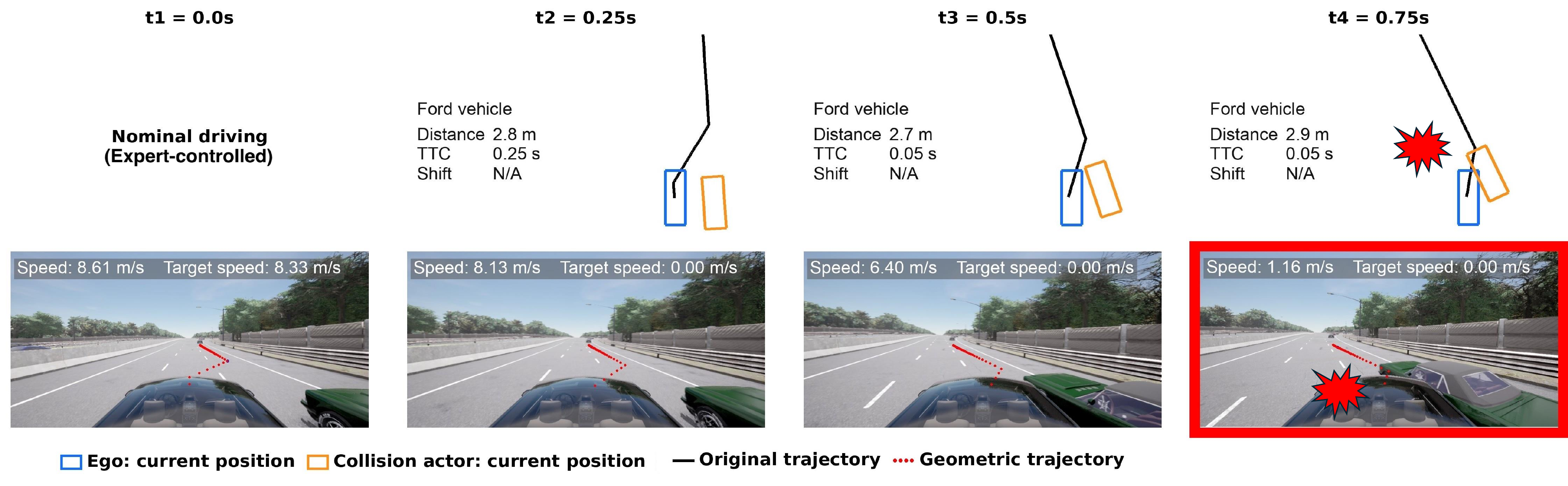}
\caption{Expert without expanded trajectory-and-speed solution space fails to prevent collision.
An expert such as PDM-Lite follows a precomputed route and primarily responds to predicted collisions with full braking along the nominal path. However, in safety-critical states, especially those induced by the policy, braking alone does not necessarily prevent an impending collision. At $t_1$, the ego vehicle controlled by expert follows the nominal route and prepares for lane change. \mbox{At $t_2$}, the expert predicts a collision and initiates full braking, but the intervention is insufficient due to the vehicle dynamics, and a \textit{collision occurs} at $t_4$. The geometric trajectories shown as red dots in the lower row correspond to the trajectories illustrated in the upper row.}
\label{fig:no_replan}
\vspace{10mm}
\centering
\includegraphics[width=2.0\columnwidth]{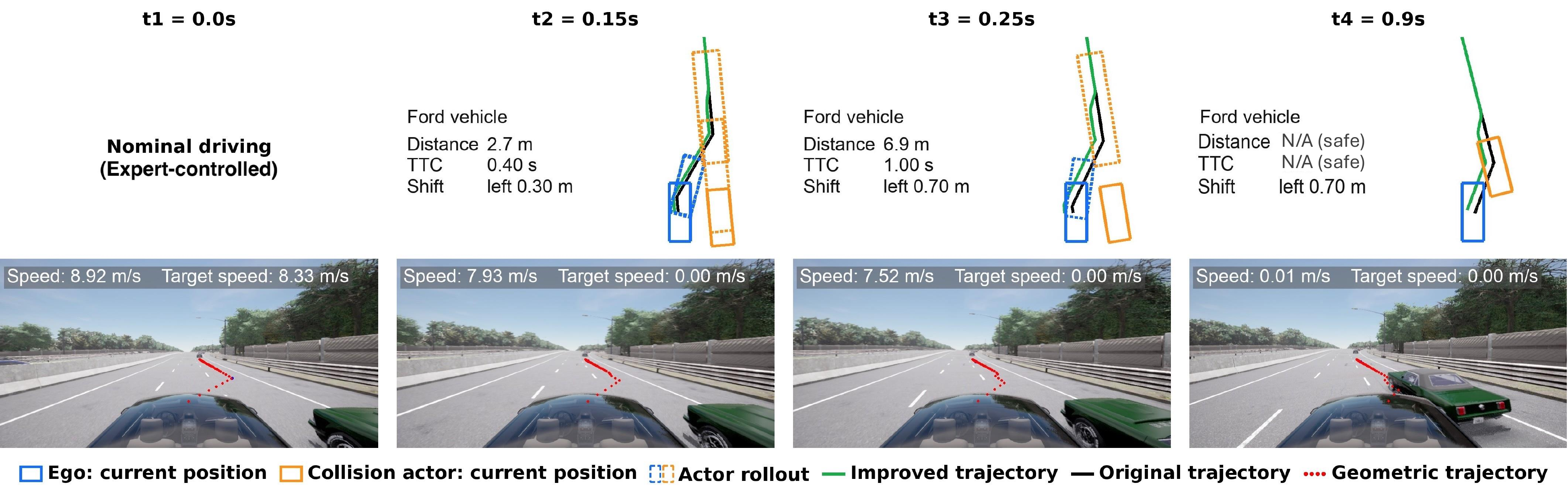}
\caption{Expert with expanded trajectory-and-speed solution space succeeds in preventing impending collision.
RoG-Expert searches for an alternative geometric trajectory when the rollout model predicts that emergency braking along the nominal trajectory is insufficient to avoid an impending collision. Specifically, at $t_1$, the ego vehicle controlled by expert follows the nominal route 
and prepares for lane change. At $t_2$, the expert proposes an emergency braking combined with a first local trajectory shift to increase clearance from the predicted collision from our rollout model. At $t_3$, the expert performs a second, larger shift as the previous plan from $t_2$ remains insufficient. The ego vehicle then continues braking safely until reaching zero speed at $t_4$. Across the two replanning steps, the predicted time-to-collision (TTC), computed by our rollout model, increases from $0.4~\mathrm{s}$ to $1.0~\mathrm{s}$, and \textit{no collision occurs}. Dashed boxes of the same color mark the first and last locations of the predicted collision actor. Note that bounding boxes are inflated following PDM-Lite to account for uncertainty in rollout prediction. The geometric trajectories shown as red dots in the lower row correspond to the improved trajectories illustrated in the upper row.}
\label{fig:replan_expert}
\end{figure*}

\subsection{Solvability-Aware Takeover Examples} \label{para:solvability_aware_examples}
\begin{figure*}
\centering
\includegraphics[width=\textwidth]{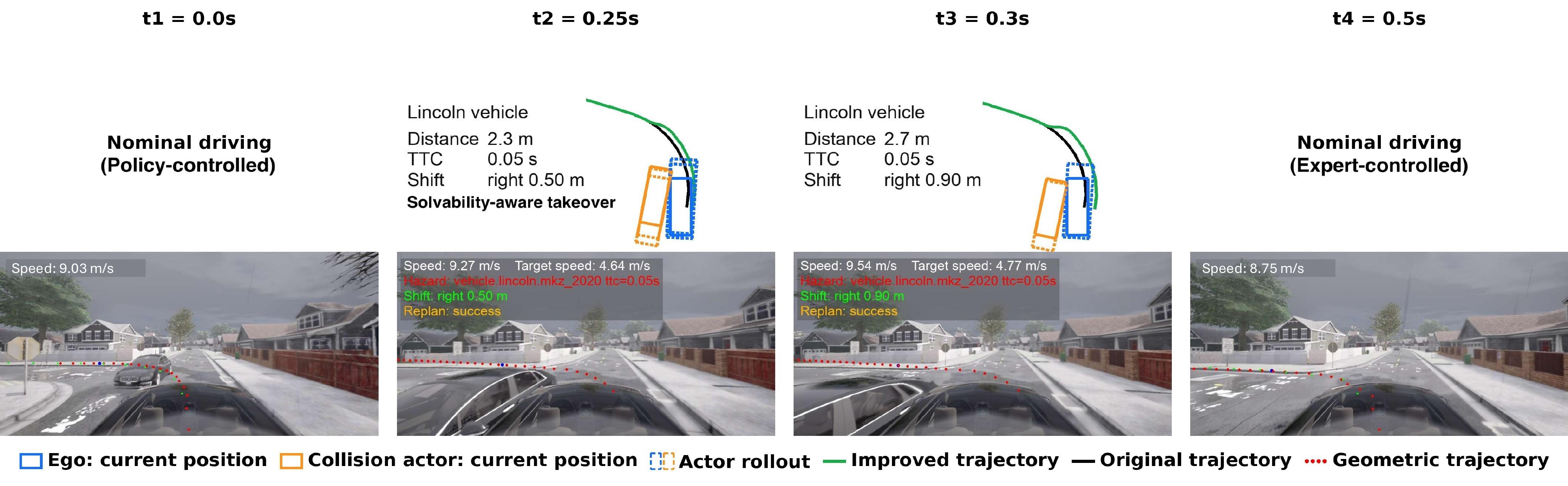}
\captionof{figure}{SimLingo with solvability-aware takeover. At $t_1$, the policy controls the ego vehicle while preparing for a left turn. As oncoming traffic approaches, our kinematic rollout model predicts at $t_2$ a time-to-collision (TTC) of $0.05\,\mathrm{s}$, indicating an imminent collision. This activates the solvability-aware takeover, which shifts the geometric trajectory $0.5\,\mathrm{m}$ to the right while setting the target speed to 50\% of the current speed. At $t_3$, the RoG-Expert applies a second, larger correction of $0.9\,\mathrm{m}$ to the right, again setting the target speed to 50\% of the current speed, successfully avoiding the collision. At $t_4$, the ego vehicle continues safely under expert control during the \mbox{post-takeover period.}}
\label{fig:solvability_takeover}
\end{figure*}

We illustrate the solvability-aware takeover mechanism in Fig.~\ref{fig:solvability_takeover}. As the ego vehicle approaches a junction, a small deviation in its turning trajectory may lead to a collision. Our solvability-aware trigger activates RoG-Expert when nominal braking becomes insufficient but corrective replanning remains feasible. RoG-Expert then adjusts the trajectory and speed to avoid the impending collision. The resulting post-takeover demonstrations provide the student with preventive supervision at safety-critical states where timely intervention is essential.

\subsection{Rule-Based Training Data Filtering}\label{data_filtering}
Following the filtering principle used in TakeAD~\cite{takead}, we retain collected samples that exhibit a meaningful discrepancy between the policy and the expert model, while removing uninformative low-motion cases. A sample is considered relevant when the target-speed disagreement exceeds 0.5 m/s, the current speed exceeds the target speed by more than 10\%, or the absolute steering deviation is at least 0.2~$\mathrm{rad}$. Speed and steering criteria are combined with OR logic, such that a disagreement in either control dimension is sufficient to retain the sample. Similarly, we filter stuck or near-stationary behavior: events with an average speed below 0.5 m/s are removed, and post-takeover segments are discarded when both their average and final speeds remain below 0.3 m/s.

In addition to these TakeAD-inspired filters, we apply an observability-based filter specific to our recovery setting. As discussed in Sec.~\ref{rog_expert}, pre-takeover collision frames are retained only when the collision-relevant actor is visible in the camera view. This prevents samples from being included when the relevant actor lies outside the model's observable field of view.

\subsection{Further Implementation Details}\label{para:further_details}
For the expert target speed, we follow LEAD~\cite{LEAD}, i.e., the target speed is capped by the minimum of the posted speed limit and the typical flow speed of nearby vehicles. In addition, we use PDM-Lite's default bounding-box inflation to account for uncertainty in both the predicted actor rollouts and the estimated execution of the ego action, unless stated otherwise. We next describe how we collected the data.
Table~\ref{tab:collection_scenarios} lists the scenario types in
fragments of CARLA Leaderboard~2.0 training routes on which SimLingo
scored below 60. None of these training routes overlap with the
Bench2Drive (B2D), Longest6 v2, or Fail2Drive \mbox{benchmark routes.}

\begin{table}[t]
\centering
\caption{\textbf{The 26 CARLA scenario types represented
in the collected route fragments.}}
\label{tab:collection_scenarios}
\resizebox{\columnwidth}{!}{%
\begin{tabular}{@{}ll@{}}
\toprule
Accident & NonSignalizedJunctionLeftTurn \\
AccidentTwoWays & NonSignalizedJunctionRightTurn \\
BlockedIntersection & OppositeVehicleRunningRedLight \\
ConstructionObstacle & OppositeVehicleTakingPriority \\
ConstructionObstacleTwoWays & ParkedObstacle \\
EnterActorFlow & ParkedObstacleTwoWays \\
HazardAtSideLane & ParkingCrossingPedestrian \\
HazardAtSideLaneTwoWays & ParkingExit \\
HighwayCutIn & PedestrianCrossing \\
HighwayExit & SignalizedJunctionLeftTurn \\
InterurbanActorFlow & SignalizedJunctionRightTurn \\
MergerIntoSlowTraffic & VehicleOpensDoorTwoWays \\
MergerIntoSlowTrafficV2 & VehicleTurningRoute \\
\bottomrule
\end{tabular}%
}
\end{table}

\section{Distribution of Post-Training Data} \label{post_trained_data}
Tab.~\ref{tab:post_training_data} summarizes the composition of the RoG-DAgger
post-training dataset from two complementary perspectives. Panel (a) reports
the proportion of samples collected before and after takeover, with the
majority originating from the post-takeover phase. Panel (b) groups samples
by their high-level collection trigger and further breaks them down into
the corresponding event types (including collisions, red-light and
out-of-lane infractions), different outcomes of the solvability-aware
takeover, and restart events.

\begin{table}[t]
    \centering
    \caption{
    \textbf{Composition of the post-training dataset from two different
    perspectives}: (a) takeover phase and (b) collection trigger and event
    type. Percentages are normalized separately within each panel.
    Bold rows in (b) denote high-level trigger classes, while indented rows
    show their \mbox{event-type breakdown}.
    }
    \label{tab:post_training_data}

    \begin{tabular*}{\columnwidth}{
        @{\extracolsep{\fill}}lr@{}
    }
        \toprule
        \multicolumn{2}{@{}l}{\textbf{(a) Takeover Phase}} \\
        \midrule
        Pre-takeover  & 23.5\% \\
        Post-takeover & 76.5\% \\

        \midrule
        \multicolumn{2}{@{}l}{
            \textbf{(b) Collection Trigger / Event Type}
        } \\
        \midrule
        \textbf{Infraction} & \textbf{51.2\%} \\
        \quad Collision & 43.9\% \\
        \quad Stop-Sign Infraction & 2.5\% \\
        \quad Red-Light Infraction & 0.2\% \\
        \quad Out-of-Lane Infraction & 4.6\% \\

        \addlinespace
        \textbf{Solvability-Aware Takeover} & \textbf{28.5\%} \\
        \quad Repeated Replan Success & 1.7\% \\
        \quad Replan Failed & 11.1\% \\
        \quad Replan Success & 15.7\% \\

        \addlinespace
        \textbf{Restart} & \textbf{20.3\%} \\
        \bottomrule
    \end{tabular*}
\end{table}
\section{Additional Ablation Studies}\label{app:add_ablation_studies}
In Sec.~\ref{para:ablation_studies}, we ablate
the three proposed components of RoG-DAgger
and pre-takeover supervision. The original ablations
use a fixed five-epoch training schedule, resulting
in fewer optimization steps when training data
is reduced. 
To rule out this confound, we run additional controls for the two ablations that reduce the training data:
removing pre-takeover samples and
removing the solvability-aware \mbox{collection run.}

As shown in Tab.~\ref{tab:add_b2d_ablation},
approximately matching the optimization steps
of the full model yields only minor improvements compared to the original ablations and
substantial performance gaps remain.
We further subsample the full RoG-DAgger dataset
to match the sample count and optimization steps
of the variant without solvability-aware collection.
Despite using the same data and training budgets,
the subsampled full model achieves a 2.41 points higher DS and 6.20~p.p.\ higher SR.
These results suggest that the performance gains
cannot be explained by data quantity or
optimization steps alone, highlighting the
importance of the 
collection strategy.

\begin{table}[t]
    \centering
    \caption{\textbf{Additional ablation studies on B2D
    to investigate the effects of optimization steps
    and training-data volume.}
    \mbox{\# Samples} includes both pre- and
    post-training data. Results are averaged over three
    evaluation runs using a single checkpoint per
    configuration. To avoid introducing additional
    data or disproportionately oversampling specific
    examples, we increase the number of training
    epochs to approximately match the full model's
    optimization steps while preserving each
    ablation's data distribution.
    }
    \label{tab:add_b2d_ablation}
    \small
    \resizebox{\columnwidth}{!}{%
    \begin{tabular}{
        @{}l
        rr
        r@{}S[table-format=-2.2]
        r@{}S[table-format=-2.2]
        @{}
    }
        \toprule
        \textbf{Ablation settings}
        & \textbf{\# Samples}
        & \makecell[c]{\textbf{\# Opt. Steps}\\
                       \textbf{(Epochs)}}
        & \multicolumn{2}{c}{\textbf{DS} $\uparrow$}
        & \multicolumn{2}{c}{\textbf{SR} $\uparrow$} \\
        \midrule

        \textbf{RoG-DAgger (full)}
        & 73,776 & 15,370 (5)
        & & 90.11
        & & 73.29 \\

        \midrule
        \quad $-$ Solvability-aware coll.
        & 43,224 & 9,005 (5)
        &  & -3.39
        &  & -7.70 \\

        \quad $-$ Pre-takeover data
        & 56,496 & 11,770 (5)
        &  & -2.35
        &  & -5.56 \\

        \midrule
        \multicolumn{7}{@{}l}{
            \textbf{Opt-step-matched (approx.)}
        } \\
        \midrule

        \quad $-$ Solvability-aware coll.
        & 43,224 & 14,408 (8)
        &  & -2.99
        &  & -6.55 \\

        \quad $-$ Pre-takeover data
        & 56,496 & 14,124 (6)
        &  & -2.15
        &  & -5.14 \\

        \midrule
        \multicolumn{7}{@{}l}{\textbf{Sample-matched}} \\
        \midrule

        \quad RoG-DAgger (subsampled)
        & 43,224 & 9,005 (5)
        &  & -0.98
        &  & -1.50 \\

        \bottomrule
    \end{tabular}%
    }
\end{table}

\section{Implementation Details of TakeVLA-noRL}\label{app:takevla_reproduction}
To benchmark RoG-DAgger against an alternative takeover and expert configuration, we implement a TakeVLA-style baseline (TakeVLA-noRL), as the official implementation is not publicly available. Specifically, we adopt the takeover triggers 
introduced
in TakeVLA and its PDM-Lite expert and integrate
them into our post-training pipeline. 
We omit TakeVLA's scenario-dreaming reinforcement learning stage, as it is orthogonal to data collection and could equally be applied on top of our approach. This isolates the comparison to the two data collection strategies.
The implementation details of the takeover triggers are provided in Tab.~\ref{tab:collection_trigger_conditions_takevla}, while the PDM-Lite expert~\cite{pdm_lite} is used as-is.

For a consistent comparison, we collect post-training data on the same CARLA LB2 routes as RoG-DAgger and use the same pretrained SimLingo-Qwen student and fine-tuning recipe. The composition of the resulting dataset is summarized in Tab.~\ref{tab:post_training_data_takevla}. Notably, TakeVLA-noRL collects approximately three times as many training samples as RoG-DAgger under the same route configuration, reflecting
differences in takeover strategies.

\begin{table}[t]
    \centering
    \caption{
       \textbf{Details on the three types of trigger conditions used for post-training data collection of TakeVLA-noRL.} Typewriter-style text refers to the original names in TakeVLA~\cite{takevla}.
    }
    \begin{tabular} {p{0.3\columnwidth} p{0.6\columnwidth}}
        \toprule
        \textbf{Trigger} & \textbf{Trigger Condition} \\
        \midrule

        \textbf{Insufficient Deceleration} \texttt{"Follow"}
        &
        A persistent discrepancy is detected
        between the ego vehicle and the ex-
        pert model: the PDM-Lite expert is braking while the student is not when the ego speed above 0.5 m/s. This has to hold for 10 consecutive steps (0.5 s).
        \\

        \midrule

        \textbf{\mbox{Constant-Velocity} TTC}
        \texttt{"Collision"}
        &
        This activates when the time-to-collision~\cite{nuplan} is under 1.0 s, assuming the ego vehicle and the other actors keep their current velocities and headings.

        \\

        \midrule

        \textbf{Restart} \texttt{"Restart"}
        &
        A persistent discrepancy is detected between the ego vehicle and the
        expert model: the ego speed remains below $0.1\,\mathrm{m/s}$ for all
        of the previous 200 frames, while both the mean expert target speed
        over these frames and the current expert target speed exceed
        $0.4\,\mathrm{m/s}$.
        \\

        \bottomrule
    \end{tabular}

    \label{tab:collection_trigger_conditions_takevla}
\end{table}

\begin{table}[t]
    \centering
    \caption{
    \textbf{Composition of the post-training dataset for TakeVLA-noRL. 
    }TakeVLA-noRL collected over 160K samples on CARLA LB2 training routes.}
    \label{tab:post_training_data_takevla}
    \begin{tabular*}{\columnwidth}{
        @{\extracolsep{\fill}}lrr@{}
    }
        \toprule
        \textbf{Collection Trigger}
        & \textbf{Frames}
        & \textbf{Share of Frames} \\
        \midrule
        Insufficient Deceleration
        & 80,363 & 47.7\% \\
        Constant-Velocity TTC
        & 59,527 & 35.3\% \\
        Restart
        & 28,596 & 17.0\% \\
        \midrule
        \textbf{Total}
        & \textbf{168,486}
        & \textbf{100.0\%} \\
        \bottomrule
    \end{tabular*}
\end{table}

\section{Additional RoG-DAgger Results}\label{add_metrics}
This section provides further details on the evaluation metrics and additional results. We first elaborate on the primary metrics used throughout the paper in App.~\ref{details_metrics}. Then, we report complementary in-distribution results on Bench2Drive, including efficiency, comfort, multi-capability performance, and additional qualitative examples, in App.~\ref{app:in_dis}. Finally, we present out-of-distribution results on Fail2Drive at both the scenario-class and individual-scenario level, together with visual examples, in App.~\ref{app:ood_eval}.
\begin{table}[t]
\centering
\caption{\textbf{Per-class harmonic mean of DS and SR on generalization scenarios of Fail2Drive.} RoG-DAgger improves over SimLingo on three out of four scenario classes, with the largest gain on behavior generalization (+30.8 HM), while SimLingo performs better on robustness scenarios.}
{
\setlength{\tabcolsep}{1.5mm}
\begin{tabular}{@{}lccr@{}}
\toprule
\textbf{Scenario Class} &
\textbf{SimLingo} &
\textbf{RoG-DAgger} &
\textbf{$\Delta$} \\
\midrule

Visual-longitudinal
& 71.1
& \textbf{81.2}
& +10.1 \\

Visual-lateral
& 45.9
& \textbf{51.3}
& +5.4 \\

Behavior
& 31.2
& \textbf{62.0}
& +30.8 \\

Robustness
& \textbf{86.8}
& 80.7
& -6.1 \\

\midrule
\textbf{Mean}
& 58.8
& \textbf{68.8}
& +10.0 \\

\bottomrule
\end{tabular}
}

\label{tab:f2d_per_class}
\end{table}
\subsection{Details of Evaluation Metrics}\label{details_metrics}
The primary evaluation metrics reported in the main paper are Route Completion (RC), Driving Score (DS), and Success Rate (SR). Route Completion measures the percentage of the prescribed route traversed by the autonomous driving agent and therefore quantifies its progress toward the destination. Driving Score jointly evaluates route progress and driving compliance by weighting the Route Completion of each route with a multiplicative infraction penalty. The penalty accounts for unsafe or invalid driving behavior, such as collisions, red-light violations, lane departures, and scenario timeouts. Success Rate measures the proportion of evaluation routes that are completed successfully while satisfying the requirements of the evaluation protocol.
\subsection{In-Distribution Evaluation}\label{app:in_dis}
In Tab.~\ref{tab:b2d_multi_ability}, we report additional B2D evaluation metrics covering driving efficiency, comfort, and multi-ability performance. RoG-DAgger generally outperforms SimLingo across these metrics and achieves performance comparable to that of TakeVLA. Further qualitative RoG-DAgger examples on B2D are presented in Fig.~\ref{fig:b2d}.

\subsection{Out-of-Distribution Evaluation}\label{app:ood_eval}

\begin{figure*}[p] 
\centering
\includegraphics[width=2.0\columnwidth]{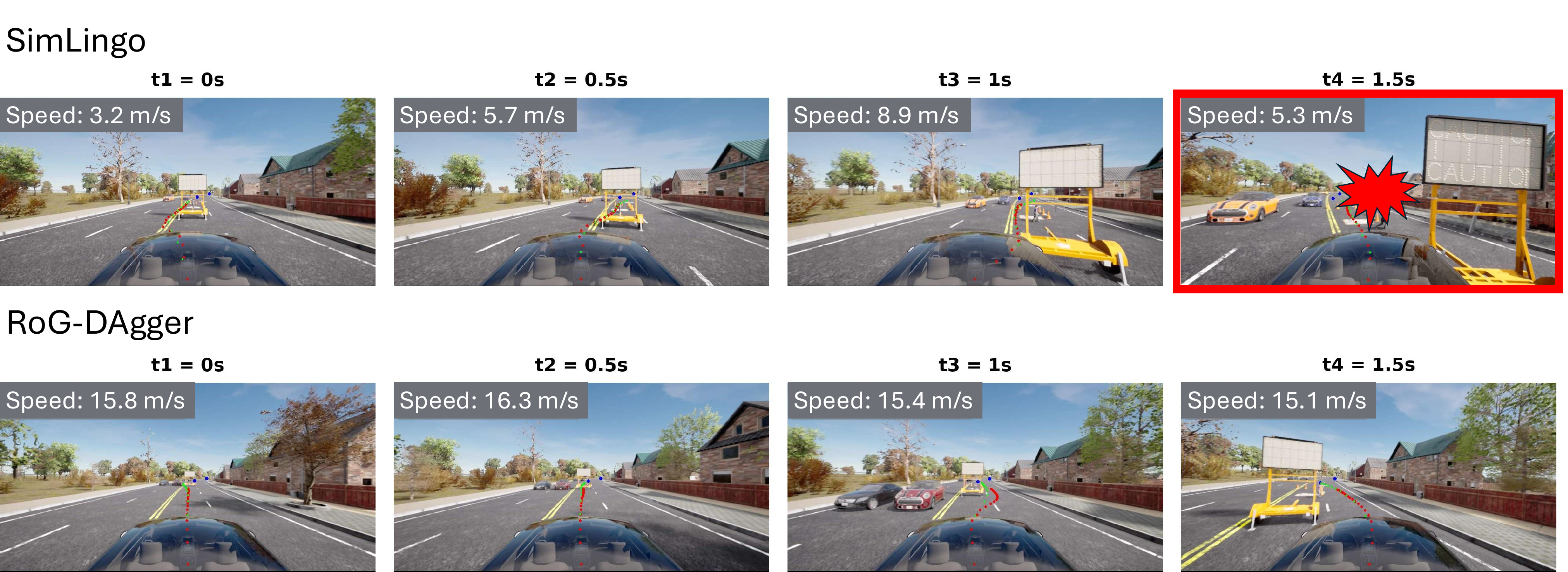}
\caption{Head-to-head comparison of RoG-DAgger (bottom) and SimLingo (top) in the \textit{Construction Pedestrian} scenario of Fail2Drive. SimLingo tends to steer left despite oncoming traffic approaching from the left while the right side of the construction site is completely free. This behavior is consistent with a training bias in which construction zones predominantly appear on the right side of the road. As a result, SimLingo may resort to a late evasive maneuver and collide with the traffic warning sign while attempting to avoid the oncoming vehicles. In contrast, RoG-DAgger learns to steer right in the same situation. This behavior can be likely attributed to the RoG-Expert, whose expanded trajectory-and-speed solution space exposes the student to multiple feasible recovery strategies and thereby encourages greater trajectory multimodality.}
\label{fig:h2h_1}
\vspace{10mm}
\centering
\includegraphics[width=2.0\columnwidth]{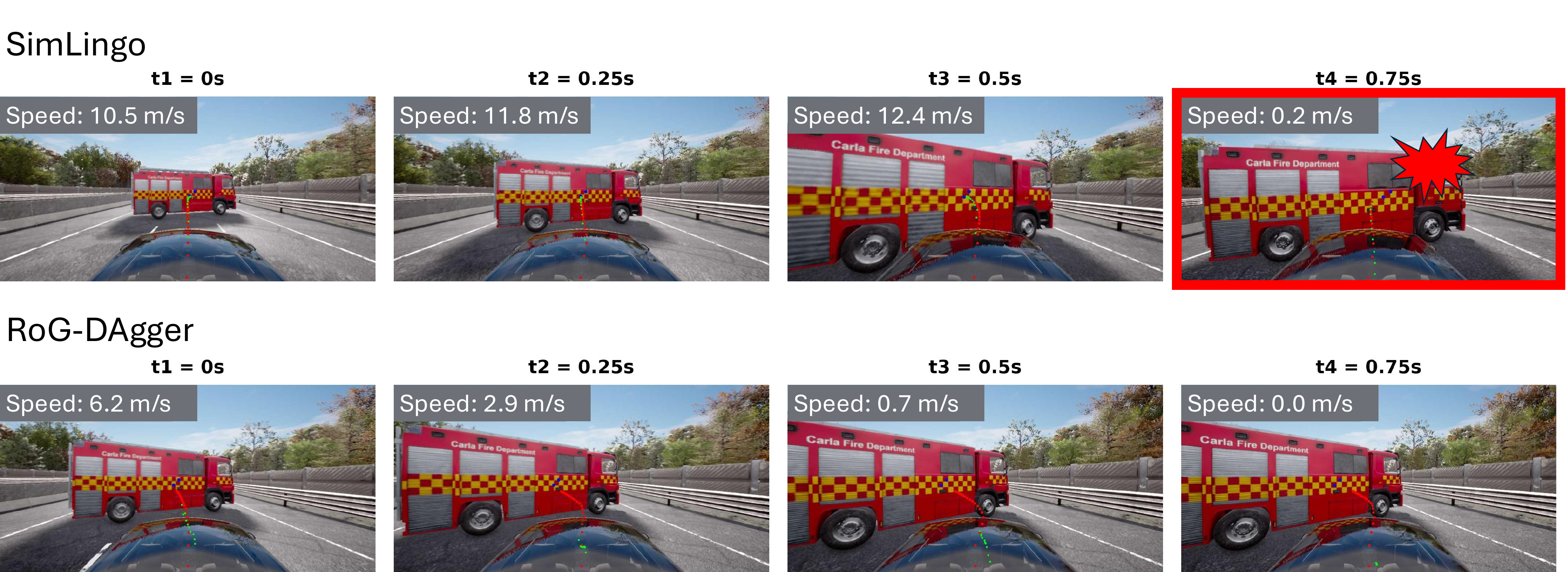}
\caption{Head-to-head comparison of RoG-DAgger (bottom) and SimLingo (top) in the \textit{Fully Blocked} scenario of Fail2Drive. SimLingo continues to accelerate from $t_1$ to $t_3$ toward the fire truck blocking the road. This behavior likely reflects a lack of corrective experience for such safety-critical situations, ultimately resulting in a collision at $t_4$. In contrast, RoG-DAgger is post-trained on safety-critical scenarios induced by the policy and paired with expert demonstrations. This preventive supervision enables the model to decelerate as it approaches the fire truck and avoid the collision.
}
\label{fig:h2h_2}
\end{figure*}

\begin{figure*}[p] 
\centering
\includegraphics[width=2.0\columnwidth]{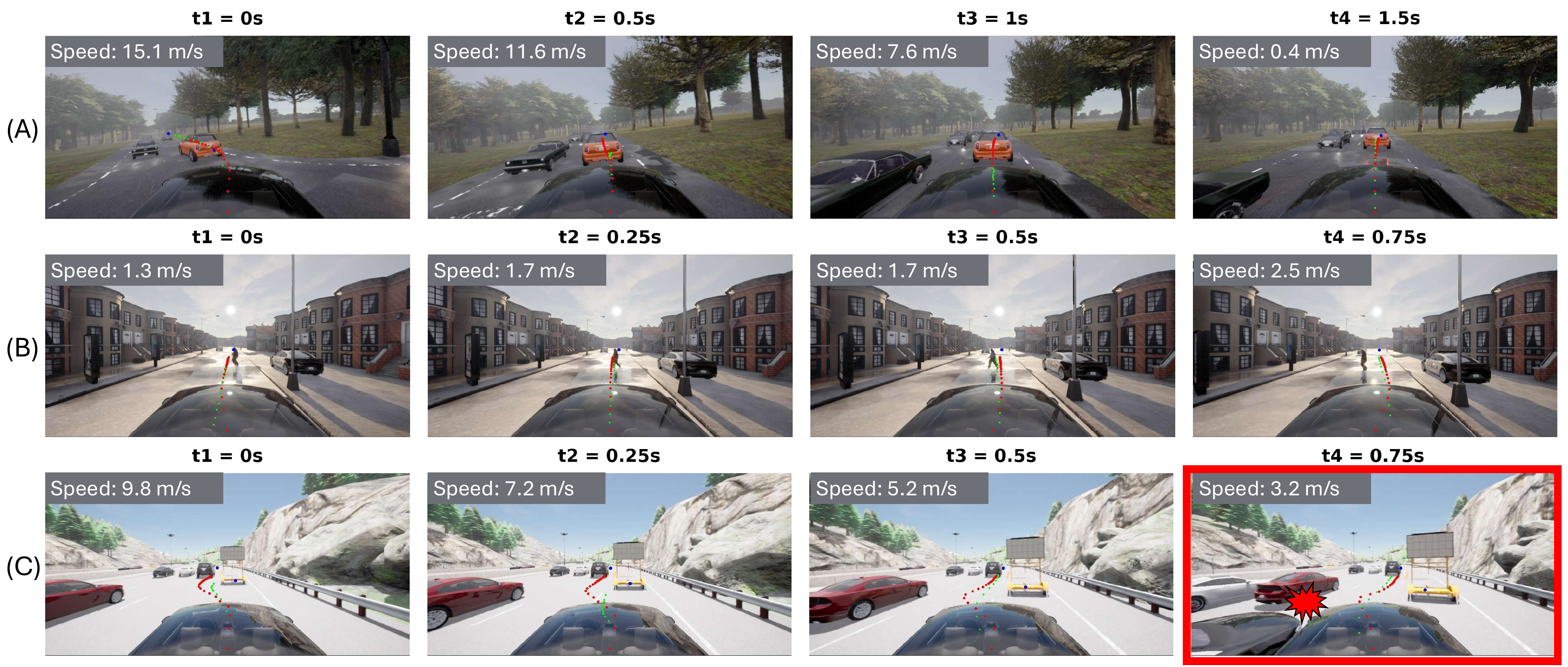}
\caption{Qualitative examples of RoG-DAgger on Bench2Drive.
In row (A), RoG-DAgger successfully merges into traffic while executing a left turn, using emergency braking to avoid a collision with the lead vehicle. In row (B), the predicted geometric trajectory adapts laterally from left to right as the pedestrian moves across the route. This contrasts with PDM-Lite/SimLingo-style behavior, where the geometric path remains largely fixed and collision avoidance is handled primarily through speed adjustment. Row (C) shows a failure case: RoG-DAgger slows down and attempts a lane change to pass through the construction area, however, due to the limited field of view, a vehicle approaching from the rear-left remains unobserved and causes a \mbox{side collision}.}
\label{fig:b2d}
\vspace{10mm}
\centering
\includegraphics[width=2.0\columnwidth]{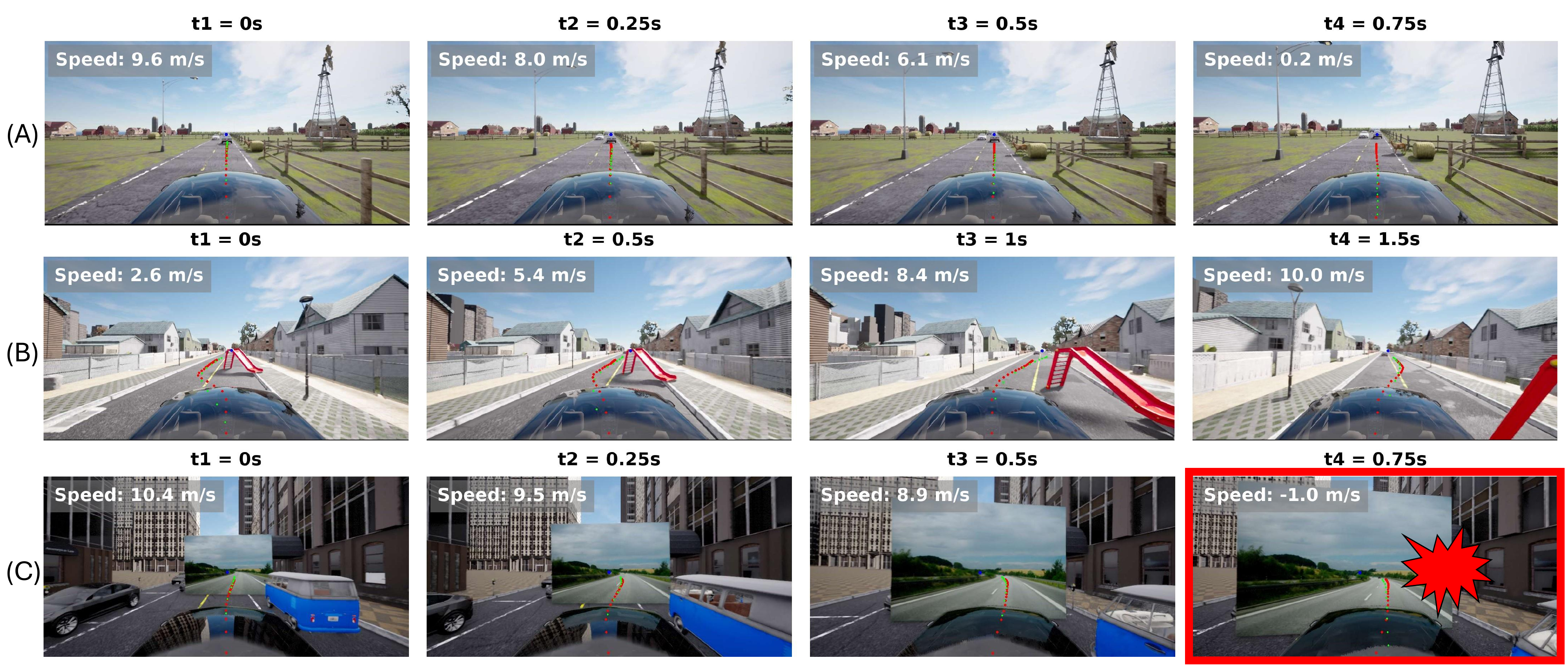}
\caption{Qualitative examples of RoG-DAgger on Fail2Drive.
In row (A), although RoG-DAgger has never been trained on scenes involving animals, the model appears to recognize that the road ahead is no longer safely drivable and begins braking in time to avoid the animal. A similar behavior is observed in row (B): despite not encountering children's slides during either pre- or post-training, the post-trained policy successfully navigates the constrained space, performs a lane change, and passes around the slide obstructing the road. Row (C) shows a failure case: Although the model appears to recognize the wall blocking the road once it is at close range, it fails to brake sufficiently and ultimately collides with the obstacle.}
\label{fig:f2d}
\end{figure*}

We report additional F2D results in Tabs.~\ref{tab:f2d_per_class} and \ref{tab:fail2drive_scenario}. Compared to SimLingo, RoG-DAgger improves harmonic mean by 30 p.p.~on \textit{Behavior} scenarios, including \textit{Fully Blocked} and \textit{Construction Pedestrian}. This gain is likely attributable to the inclusion of a broader range of safety-critical scenarios during post-training. In contrast, RoG-DAgger performs 6 p.p.~worse on \textit{Robustness} scenarios. We hypothesize that the model exhibits overly conservative behavior near construction sites with irregular object orientations and placements, which can impede route progress and lead to scenario-timeout penalties. We provide additional qualitative head-to-head comparisons of RoG-DAgger and SimLingo in Figs.~\ref{fig:h2h_1} and~\ref{fig:h2h_2}. Further qualitative examples for F2D are presented in Fig.~\ref{fig:f2d}.

\section{Technical Details}
We conduct our experiments on Microsoft Azure Standard\_NC96ads\_A100\_v4 instances, equipped with 96 AMD EPYC 7V13 (Milan) CPU cores, 880 GB of system memory, and four NVIDIA A100 PCIe GPUs with 80 GB of memory each. Our software environment uses Python 3.10, PyTorch 2.8 with CUDA 12.6 and cuDNN 9.10, and CARLA 0.9.15 for closed-loop simulation.
Using four NVIDIA A100 GPUs, pre-training SimLingo-Qwen3VL-2B takes approximately 35 hours with our training schedule. Data collection on CARLA Leaderboard 2.0 requires approximately 15 hours with six parallel CARLA instances per GPU, and post-training takes an additional \mbox{3-4 hours}.

\begin{table*}[p]
\centering

{
\setlength{\tabcolsep}{1mm}
\caption{\textbf{Detailed Bench2Drive evaluation of efficiency, comfort, and multi-ability performance.} RoG-DAgger improves over SimLingo in merging, overtaking, and traffic-sign handling, and achieves a higher overall mean. It performs comparably to TakeVLA overall, with stronger merging but lower emergency-braking performance. Beyond these scenario-specific capabilities, we further assess whether the observed gains preserve desirable overall driving behavior in terms of efficiency and comfort. RoG-DAgger maintains competitive efficiency while improving comfort over both SimLingo and TakeVLA, suggesting that its closed-loop gains do not come at the cost of less comfortable driving.}
\label{tab:b2d_multi_ability}
\begin{tabular}{@{}lccccccc>{\columncolor{gray!15}}c@{}}
\toprule
\multirow[c]{2}{*}{\textbf{Method}} &
\multirow[c]{2}{*}{\textbf{Efficiency}} &
\multirow[c]{2}{*}{\textbf{Comfort}} &
\multicolumn{6}{c}{\textbf{Multi-Capability}~(\%)} \\
\cmidrule(lr){4-9}
& & &
\textbf{Merging} &
\textbf{Overtaking} &
\makecell[c]{\textbf{Emergency}\\\textbf{Brake}} &
\makecell[c]{\textbf{Give}\\\textbf{Way}} &
\makecell[c]{\textbf{Traffic}\\\textbf{Sign}} &
\cellcolor{white}\textbf{Mean} \\
\midrule

\multicolumn{9}{l}{\hspace{1em}\textit{Privileged methods}} \\
\midrule

PDM-Lite~\cite{pdm_lite}
& 213.47
& 22.33
& 82.05
& 91.11
& 81.67
& 80.00
& 70.97
& 81.16 \\

RoG-Expert~\textbf{(ours)}
& 196.89
& 14.13
& 75.00
& 84.44
& 93.33
& 40.00
& 88.95
& 76.35 \\

\midrule
\multicolumn{9}{l}{\hspace{1em}\textit{Pre-trained-only methods}} \\
\midrule

UniAD~\cite{uniad}
& 129.21
& 43.58
& 12.16
& 20.00
& 23.64
& 10.00
& 13.89
& 15.89 \\

TF$++$~\cite{pdm_lite}
& --
& --
& 58.75
& 57.77
& 83.33
& 40.00
& 82.11
& 64.39 \\

ORION~\cite{orion}
& 151.48
& 17.38
& 25.00
& 71.11
& 78.33
& 30.00
& 69.15
& 54.72 \\

HiP-AD~\cite{hipad}
& 203.12
& 19.36
& 50.00
& 84.44
& 83.33
& 40.00
& 72.10
& 65.98 \\

AutoVLA~\cite{autovla}
& 146.93
& 39.33
& --
& --
& --
& --
& --
& -- \\

SimLingo~\cite{simlingo}
& 259.23
& 33.67
& 54.01
& 57.04
& 88.33
& 53.33
& 82.45
& 67.03 \\

\midrule
\multicolumn{9}{l}{\hspace{1em}\textit{Post-trained methods}} \\
\midrule

MindDrive~\cite{minddrive}
& --
& --
& 32.89
& \textbf{75.56}
& 68.33
& 50.00
& 57.89
& 56.94 \\

TakeAD~\cite{takead}
& 193.30
& 22.89
& 30.77
& 35.56
& 56.67
& 50.00
& 42.02
& 43.00 \\

TakeVLA~\cite{takevla}
& \textbf{249.01}
& 30.27
& 63.64
& 64.44
& \textbf{91.67}
& \textbf{50.00}
& \textbf{85.48}
& \textbf{71.05} \\

RoG-DAgger~\textbf{(ours)}
& 242.84
& \textbf{35.10}
& \textbf{67.09}
& 64.44
& 85.00
& \textbf{50.00}
& 84.74
& 70.25 \\

\bottomrule
\end{tabular}
}
\vspace{3mm}
\centering

{
\setlength{\tabcolsep}{1mm}
\caption{\textbf{Per-scenario performance of RoG-DAgger on Fail2Drive.} We report Driving Score (DS) and Success Rate (SR) on the \textit{Base} and \textit{Generalization} splits. SR and relative changes between the two splits (Rel.) are reported as percentages. RoG-DAgger performs particularly well on several novel scenario configurations, such as \textit{Construction Pedestrian} and \textit{Fully Blocked}. However, it still fails on entirely novel scenarios involving unseen assets, such as \textit{Wall}.}
\label{tab:fail2drive_scenario}
\vspace{4pt}
\begin{tabular}{@{}llrr>{\columncolor{gray!15}}rrr>{\columncolor{gray!15}}r@{}}
\toprule
\multirow{2}{*}{\textbf{Class}} &
\multirow{2}{*}{\textbf{Scenario}} &
\multicolumn{3}{c}{\textbf{Driving Score (DS)} $\uparrow$} &
\multicolumn{3}{c}{\textbf{Success Rate (SR)} $\uparrow$} \\
\cmidrule(lr){3-5}
\cmidrule(lr){6-8}
&
&
\textbf{Base} &
\textbf{Gen.} &
\cellcolor{white}\textbf{Rel.} &
\textbf{Base} &
\textbf{Gen.} &
\cellcolor{white}\textbf{Rel.} \\
\midrule

Behavior
& Wall
& 98.6 & 68.5 & -30.5
& 100.0 & 20.0 & -80.0 \\

Behavior
& Fully Blocked
& 94.1 & 83.8 & -11.0
& 100.0 & 80.0 & -20.0 \\

Behavior
& Pedestrians On Road
& 82.6 & 64.8 & -21.5
& 80.0 & 60.0 & -25.0 \\

Behavior
& Construction Pedestrian
& 54.1 & 67.4 & +24.6
& 40.0 & 60.0 & +50.0 \\

\midrule

Visual-lat
& Bad Parking
& 79.3 & 71.0 & -10.5
& 70.0 & 60.0 & -14.3 \\

Visual-lat
& Construction Permutations
& 76.4 & 54.3 & -28.9
& 100.0 & 40.0 & -60.0 \\

Visual-lat
& Custom Obstacles
& 51.8 & 55.6 & +7.2
& 30.0 & 30.0 & 0.0 \\

\midrule

Visual-lon
& Obscured Stop
& 95.4 & 99.4 & +4.2
& 100.0 & 100.0 & 0.0 \\

Visual-lon
& Hard Brake
& 83.3 & 73.8 & -11.4
& 80.0 & 60.0 & -25.0 \\

Visual-lon
& Right Of Way
& 90.9 & 83.7 & -8.0
& 100.0 & 80.0 & -20.0 \\

Visual-lon
& Animals
& 78.6 & 82.1 & +4.6
& 70.0 & 80.0 & +14.3 \\

Visual-lon
& Pedestrian Other Blocker
& 73.8 & 74.1 & +0.4
& 80.0 & 80.0 & 0.0 \\

\midrule

Robustness
& Right Construction
& 93.7 & 72.5 & -22.7
& 100.0 & 60.0 & -40.0 \\

Robustness
& Opposite Construction
& 98.5 & 98.6 & +0.2
& 100.0 & 100.0 & 0.0 \\

Robustness
& Image On Object
& 79.9 & 73.5 & -8.0
& 100.0 & 80.0 & -20.0 \\

Robustness
& Passable Obstacles
& 92.5 & 97.5 & +5.3
& 100.0 & 100.0 & 0.0 \\

Robustness
& Pedestrian Crowd
& 71.8 & 64.5 & -10.2
& 80.0 & 60.0 & -25.0 \\

\bottomrule
\end{tabular}
}

\end{table*}

